\documentclass{article}

\PassOptionsToPackage{numbers, compress}{natbib}

\usepackage[final]{neurips_2022}

\usepackage[utf8]{inputenc} 
\usepackage[T1]{fontenc}    
\usepackage{hyperref}       
\usepackage{url}            
\usepackage{booktabs}       
\usepackage{amsfonts}       
\usepackage{nicefrac}       
\usepackage{microtype}      
\usepackage{xcolor}         
\usepackage{amsmath}
\usepackage{graphicx}
\usepackage{amsthm}
\usepackage{tabu}
\usepackage{subfig}
\usepackage{multirow}
\usepackage{textcomp}
\usepackage{pifont}

\newcommand{\minisection}[1]{\vspace{2mm}\noindent{\textbf{#1}.}}

\newcommand{\Sref}[1]{Sec.~\ref{#1}}

\newcommand{\Fref}[1]{Fig.~\ref{#1}}
\newcommand{\Tref}[1]{Table~\ref{#1}}

\DeclareMathOperator*{\argmin}{argmin}

\newcommand{\fourimgperrow}[1]{
    \includegraphics[width=3.2cm, height=3.2cm]{#1}
}
\newcommand{\threeimgperrow}[1]{
    \includegraphics[width=4cm, height=4cm]{#1}
}

\title{QC-StyleGAN - Quality Controllable Image Generation and Manipulation}

\author{%
Dat Viet Thanh Nguyen$^{1,}$\thanks{authors contributed equally}  \quad\qquad Phong Tran$^{1,2,}$\footnotemark[1] \quad\qquad Tan M. Dinh$^{1}$ \vspace{8pt}\\
\textbf{Anh Tuan Tran$^{1}$} \quad\qquad
\textbf{Cuong Pham$^{1,3}$} \vspace{8pt} \\
\small{\textsuperscript{1}VinAI Research \quad \textsuperscript{2}MBZUAI \quad \textsuperscript{3}Posts \& Telecommunications Institute of Technology} \vspace{3pt}\\
\texttt{\scriptsize \{v.datnvt2, v.tandm3, v.anhtt152, v.cuongpv11\}@vinai.io} \quad \texttt{\scriptsize the.tran@mbzuai.ac.ae}
}

\begin{document}

\maketitle

\begin{abstract}
The introduction of high-quality image generation models, particularly the StyleGAN family, provides a powerful tool to synthesize and manipulate images. However, existing models are built upon high-quality (HQ) data as desired outputs, making them unfit for in-the-wild low-quality (LQ) images, which are common inputs for manipulation. In this work, we bridge this gap by proposing a novel GAN structure that allows for generating images with controllable quality. The network can synthesize various image degradation and restore the sharp image via a quality control code. Our proposed QC-StyleGAN can directly edit LQ images without altering their quality by applying GAN inversion and manipulation techniques. It also provides for free an image restoration solution that can handle various degradations, including noise, blur, compression artifacts, and their mixtures. Finally, we demonstrate numerous other applications such as image degradation synthesis, transfer, and interpolation. The code is available at \url{https://github.com/VinAIResearch/QC-StyleGAN}.
\end{abstract}

\section{Introduction}
Image generation has achieved a marvelous development in recent years thanks to the introduction of Generative Adversarial Networks (GAN). StyleGAN models \cite{karras2019stylegan,Karras2019stylegan2,karras2020ada,karras2021alias} manage to generate realistic-looking images with the resolution up to 1024$\times$1024. Their synthetic images of non-existing people/objects can fool human eyes \cite{ThisXNoExist,WhichFaceReal}. 
The development of such high-quality synthesis models has also introduced  a new direction to effectively solve the image manipulation tasks, which usually 
first fit the input image to the model's latent space via a GAN ``inversion'' technique \cite{abdal2019image2stylegan,zhu2020domain,richardson2021encoding,tov2021designing,alaluf2021restyle,alaluf2021hyperstyle,dinh2021hyperinverter}, then apply a learned editing \cite{shen2020interfacegan,harkonen2020ganspace,patashnik2021styleclip} on the fitted latent code for the desired change on the generated image. 
Their impressive manipulation results promise various practical applications in entertainment, design, art, and more.

While the recent GAN models, notably the StyleGAN series, show promising image editing performance, we argue that image quality is an Achilles' heel, making it challenging for them to work on real-world images. StyleGAN models are often trained from sharp and high-resolution images as the desired output quality. In contrast, in-the-wild images might have various qualities depending on the capturing and storing conditions. Many degradations could affect these images, including noises, blur, downsampling, or compression artifacts. They make GAN inversion hard, sometimes impossible, to fit the low-quality inputs to the high-quality image domain modelled by StyleGAN generators. Incorrect inversions might lead to unsatisfactory editing results with obvious content mismatches. For example, a popular StyleGAN-based image super-resolution method \cite{menon2020pulse} caused a controversy by approximating a high-resolution picture of Barack Obama as an image of a white man.

One possible solution to narrow the quality gap is to train the StyleGAN generator on low-quality images. Although this might improve GAN inversion accuracy, it can possibly fail to model the high-quality image distribution, which is the desired target of standard image generation and many image manipulation tasks. Training the generator on mixed quality data also does not help since the connection to translate between low-quality and high-quality images is missing.

This paper resolves the aforementioned problems by introducing a novel \textbf{\underline{Q}}uality \textbf{\underline{C}}ontrollable \textbf{\underline{StyleGAN}} structure, or \textbf{QC-StyleGAN}, which is a simple yet effective architecture that can learn and represent a wide range of image degradations and encode them in a controllable vector. We demonstrate our solution by modifying StyleGAN2-Ada networks but it should be applicable to any StyleGAN version. Based on the standard structure, we revise its fine-level layers to input a quality code $q$ defining the degradations on the model output. It can generate clean and sharp images similar to the standard StyleGAN counterpart when $q = 0$ and synthesize their degraded, low-quality versions by varying the value $q$. Our QC-StyleGAN covers popular degradations, including noise, blur, low-resolution and downsampling, JPEG compression artifacts, and their mixtures. It has many desired properties. First, QC-StyleGAN can generate sharp and clean images with almost the same quality as the standard StyleGAN, thus preserving all of its utilities and applications. Second, it can also model the degraded photos, bridging the gap to real-world images and providing more accurate GAN inversion. The image editing operators, therefore, can be applied on low-quality images in the same way as for the high-quality ones without altering image quality. This functionality is particularly important when only a part of the data is manipulated, and the manipulated result must have consistent quality as the rest. One scenario is to edit a few frames in a video. Another scenario is to edit an image crop of a big picture. Third, it allows easy conversion between low- and high-quality outputs, bringing in many applications below.

\begin{figure}[t]
\centering
\includegraphics[width=0.7\linewidth]{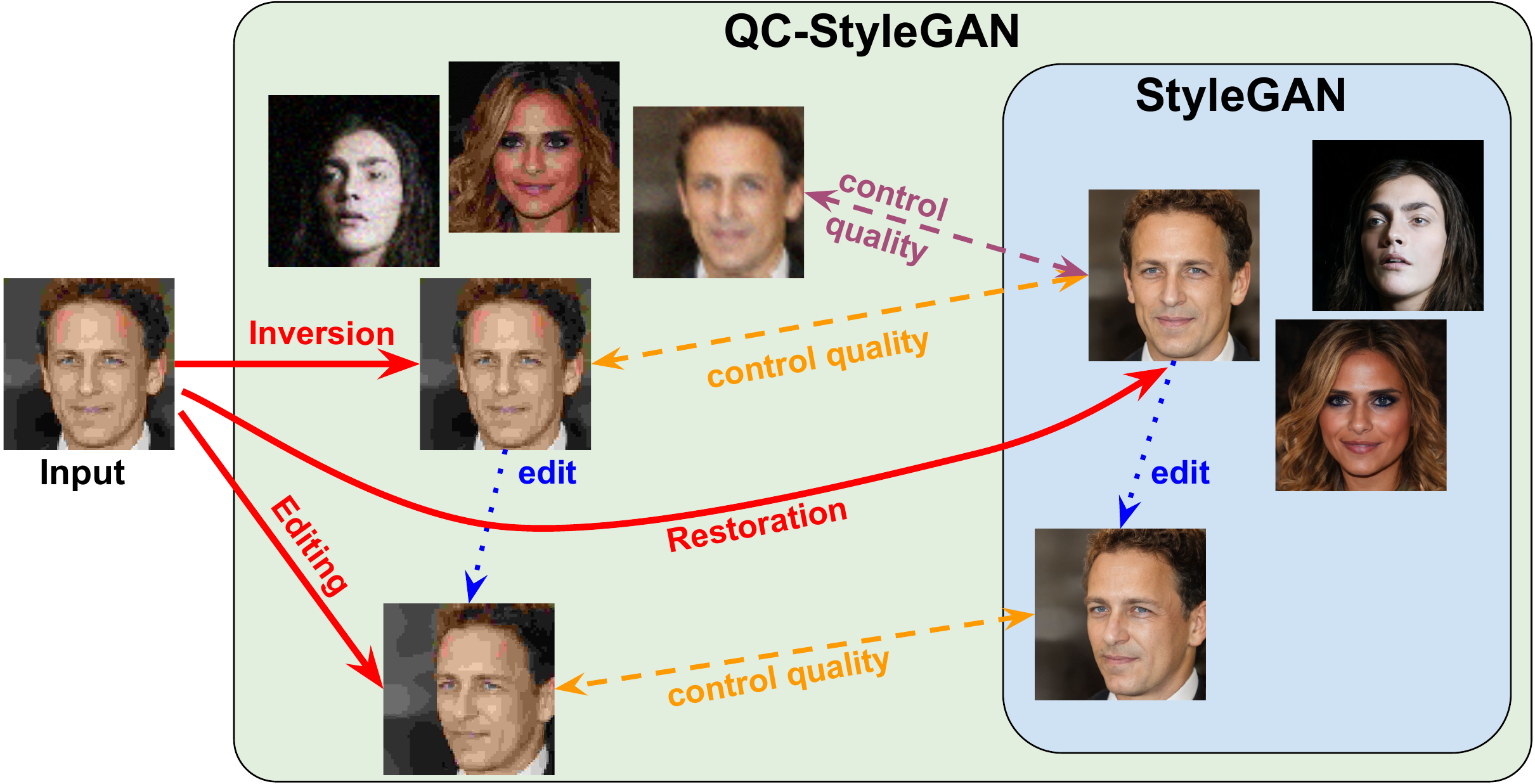}
\caption{Our QC-StyleGAN allows for synthesizing sharp images, similar to the standard StyleGAN, and degraded images. It provides a quality-control input for easy conversion between degraded images and their sharp versions (dashed arrows). The same quality codes produce the same degradation (yellow arrows), and QC-StyleGAN covers a wide range of degradations (yellow vs. magenta). We can easily edit degraded images using the editing directions learned for sharp images in StyleGAN space (blue dotted arrows). Given a low-quality input, QC-StyleGAN allows more accurate GAN inversion, direct image editing with quality preserved, and an efficient image restoration (red arrows).}
\vspace{-5mm}
\label{fig:teaser}
\end{figure}

One of our QC-StyleGAN's applications is GAN-based image restoration. Given a low-quality input image, we can fit it into QC-StyleGAN, then recover to the high-quality, sharp version by resetting the quality code $q$ to 0. While there were several works that used the StyleGAN prior for image super-resolution \cite{menon2020pulse,chan2021glean} or deblurring \cite{tran2021enblur}, QC-StyleGAN is the first method to solve the general image restoration task. Moreover, while the previous methods tried to fit low-quality data to the high-quality image space, leading to obvious content mismatches, our model maintains the content consistency by bridging the two data realms. This image restoration, while being a simple by-product of our QC-StyleGAN design, shows great potential in handling images with complex degradations.  

Besides, by modeling a wide range of image degradations and encoding them in a controllable vector, QC-StyleGAN can be used to synthesize novel image degradations or interpolate between existing ones. It can easily capture the degradation from an input image, allowing degradation transfer. These techniques have various practical applications and will be demonstrated in Section \ref{sec:exp}.

Fig. \ref{fig:teaser} summarizes our proposed QC-StyleGAN. It covers not only the sharp, high-quality image domain of common StyleGAN models but also the degraded, low-quality image one. It bridges these image domains via a quality control input. QC-StyleGAN allows better GAN inversion and image editing on low-quality inputs and introduces a potential image restoration method.



\section{Related work}
\subsection{StyleGAN series} 
Since the seminal paper \cite{goodfellow2014generative}, Generative Adversarial Networks (GANs) have achieved tremendous progress. Among them, a typical line of work called Style-based GANs (StyleGANs) has attracted much attention from the research community. These works allow us to generate images at very high resolution while producing semantically disentangled and smooth latent space. In the initial version \cite{karras2019stylegan}, StyleGAN controls the style of synthesized images by proposing an intermediate disentangled latent space, named $\mathcal{W}$ space, mapped from the latent code via an MLP network. Then, they feed this code to the generator at each layer by employing the adaptive instance normalization module. In the next generation, StyleGAN2~\cite{Karras2019stylegan2} proposed a few changes in the network design and training components 
to further enhance the image quality. StyleGAN-Ada~\cite{karras2020ada} allows model training with limited data by introducing the adaptive discriminator augmentation technique.
StyleGAN3~\cite{karras2021alias} tackles the aliasing artifact phenomenon in the previous versions and therefore helps to generate images entirely equivariant for rotation and translation. Recently, StyleGAN-XL~\cite{Sauer2021ARXIV} expanded the ability of the StyleGAN model to synthesize images on the ImageNet~\cite{krizhevsky2012imagenet} dataset. It is worth noting that all of the current StyleGAN models have used the training dataset with high-quality images. Therefore, their outputs also are sharp images. In our work, we explore a new StyleGAN structure that allows us to synthesize both high- and low-quality images with explicitly quality control.

\subsection{Latent space traversal and GAN inversion}\label{sec:rel_inversion} 
Aside from the ability to synthesize high-quality images, the latent space learned by GANs also encodes a diverse set of interpretable semantics, making it an excellent tool for image manipulation. As a result, exploring and controlling the latent space of GANs has been the focus of numerous research works. Many studies~\cite{goetschalckx2019ganalyze,shen2020interpreting, yang2021discovering} have tried to extract the editing directions from the latent space in a supervised manner by leveraging either a pre-trained attribute classifier or a set of attribute-annotated images. Meanwhile, other works such as ~\cite{harkonen2020ganspace,voynov2020unsupervised, wang2020geometric,shen2021closedform} have developed the unsupervised methods for mining the latent space, which reveal many new interesting editing directions. To convey such benefits for editing real images, we first need to obtain the latent code in the latent space so that we can accurately reconstruct the input image when we feed this code into the pre-trained generator. This line of work is called \emph{GAN inversion}, which was first proposed by~\cite{zhu2016generative}. Existing GAN inversion techniques can be grouped into (1) optimization-based~\cite{creswell2018inverting,ma2019invertibility,abdal2019image2stylegan,Karras2019stylegan2,kang2021gan}; (2) encoder-based~\cite{zhu2016generative,pidhorskyi2020adversarial,richardson2021encoding,tov2021designing,wei2021simpleinversion} and (3) two-stage ~\cite{bau2019seeing,zhu2020domain,guan2020collaborative,bau2020semantic,tewari2020pie,roich2021pivotal,dinh2021hyperinverter,  alaluf2021hyperstyle,wang2021high} approaches. We recommend visiting the comprehensive survey~\cite{xia2021survey} for a more in-depth review.

\subsection{Image enhancement and restoration}
Image enhancement and restoration are the task of increasing the quality of a given degraded image. Formally, the degradation process can be generalized as:
\begin{align}
    y = \mathcal{H}(x) + \eta
    \label{eq:restoration}
\end{align}
where $\mathcal{H}$ is the degradation operator, $\eta$ is noise, $x$ and $y$ are the original and the degraded images, respectively. The goal is to find $x$ given $y$ and probably some assumptions on $\mathcal{H}$ and $\eta$. It can be divided into various sub-tasks based on the degraded operator and the noise such as image denoising ($\mathcal{H} = I$), image deblurring ($\mathcal{H}$ is a blur operator), or image super-resolution ($\mathcal{H}$ is a downsample operator). Existing methods usually focus on one of these sub-tasks instead of solving the general one. Image enhancement and restoration is a well-studied yet still challenging field. In the past, common methods make handcrafted priors on $\mathcal{H}$ and $\eta$ and use complicated optimization algorithms to solve $x$ \cite{he2010single,pan2016blind,dabov2007image,tsai1989multiple}. Recently, many deep-learning-based methods have been proposed and achieved impressive results. These methods mainly differ by the network design \cite{dong2015image,zhang2017beyond,tao2018scale}. However, data-driven approaches were observed to be highly overfitted to the training set and hence cannot be applied for real-world degraded images. 

To better restore in-the-wild images, recent works consider the task on a specific domain, such as face \cite{menon2020pulse,chan2021glean,tran2021explore}, by leveraging existing generative models, such as StyleGAN. Unlike previous deep-based models, these methods always produce high-quality results even on real-world degraded images. However, they often produce clear mismatched image content when trying to fit the degraded input into the sharp image space.

\section{Proposed method}\label{sec:method}
In this section, we present our proposed QC-StyleGAN that supports quality-controllable image generation and manipulation. We first define the QC-StyleGAN concept (Section \ref{sec:overview}), then discuss its structure (Section \ref{sec:structure}) and training scheme (Section \ref{sec:loss}). Next, we discuss the technique to acquire precise inversion results (Section \ref{sec:inversion}). Finally, we present its various applications (Section \ref{sec:applications}).

\subsection{Problem definition}\label{sec:overview}
Traditional image generators input a random noise vector $z \in \mathbb{R}^{D_i} \sim \mathcal{N}(0, I)$, with $D_i$ as the number of input dimensions, and output the corresponding sharp image. Let us denote the baseline StyleGAN generator as $\mathcal{F}_0$. The image generation process is $I = \mathcal{F}_0(z)$. Furthermore, $\mathcal{F}_0$ consists of two components: a mapping network $\mathcal{M}_0$ and a synthesis network $\mathcal{G}_0$:
\begin{align}
  \mathcal{F}_0 = \mathcal{G}_0 \circ \mathcal{M}_0, \qquad \qquad I = \mathcal{F}_0(z) = \mathcal{G}_0(\mathcal{M}_0(z)) = \mathcal{G}_0(w),
\end{align}
with $w = \mathcal{M}_0(z)$ forming a commonly used embedding space $\mathcal{W}$. 

Our proposed network, denoted as $\mathcal{F}$, requires an extra quality code input, denoted as $q \in \mathbb{R}^{D_q} \sim \mathcal{N}(0, I)$, with $D_q$ as the number of quality-code dimensions. Its image generation process is $I = \mathcal{F}(z, q)$. We borrow the mapping function $\mathcal{M}_0$ and design a new synthesis network $\mathcal{G}$ for $\mathcal{F}$:
\begin{align}
  \mathcal{F} = \mathcal{G} \circ \mathcal{M}_0, \qquad \qquad I = \mathcal{F}(z, q) = \mathcal{G}(\mathcal{M}_0(z), q) = \mathcal{G}(w, q).
\end{align}
The desired network should satisfy two requirements:
\begin{enumerate}
    \item When $q$ is zero, $\mathcal{F}$ synthesizes sharp images similar to the standard StyleGAN:
    \begin{equation}
        \mathcal{F}(z, 0) = \mathcal{F}_0(z).
    \end{equation}
\item  When $q$ is nonzero, $\mathcal{F}$ synthesizes a degraded version of the corresponding sharp image:
    \begin{equation}
        \mathcal{F}(z, q) = \mathcal{A}_q(\mathcal{F}_0(z)),
    \label{eq:im_degraded}
    \end{equation}
    where $\mathcal{A}_q$ is some image degradation function with the parameter $q$. $\mathcal{A}_q$ is a composition of multiple primitive degradation functions such as noise, blur, image compression, and more.
    It is defined by $q$ and independent to the sharp image content $\mathcal{F}_0(z)$. 
\end{enumerate}


\subsection{Network structure}\label{sec:structure}
\begin{figure}[t]
\centering
\subfloat[Network structure]{
\includegraphics[width=.355\textwidth]{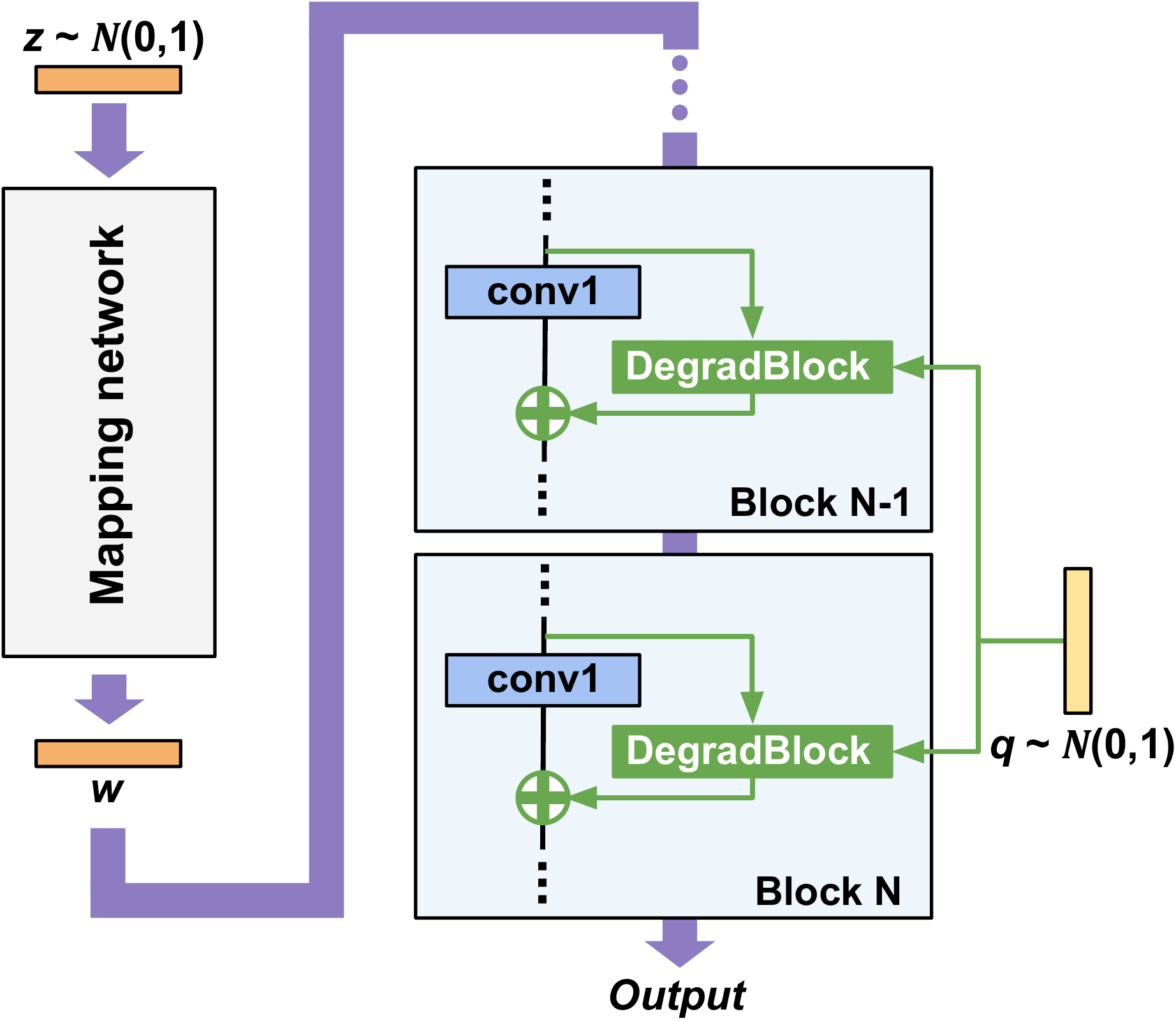}
\label{fig:structure}
}
\hspace{3mm}
\subfloat[DegradBlock]{
\includegraphics[width=.51\textwidth]{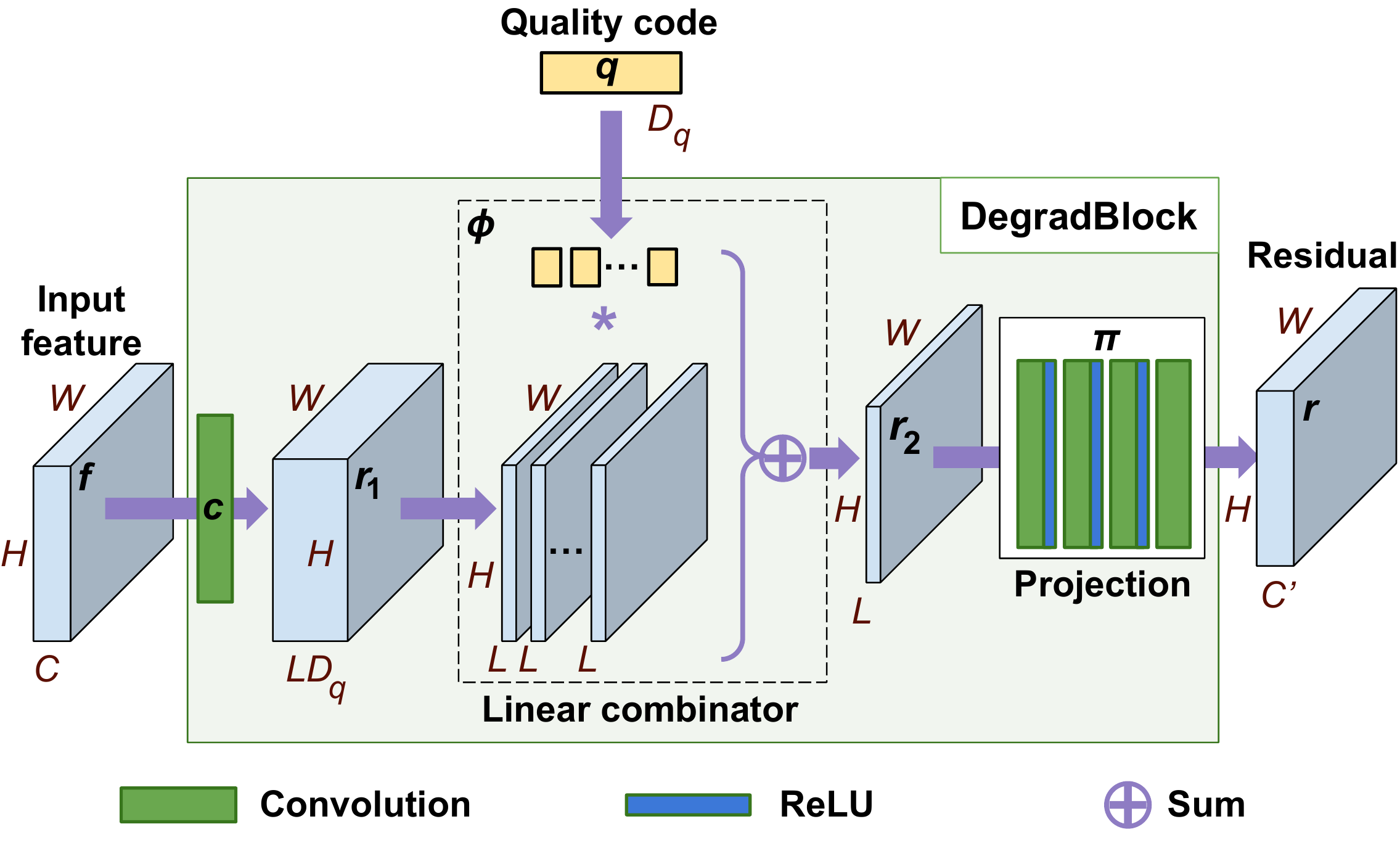}
\label{fig:degradblock}
    }
\vspace{-2mm}
\caption{QC-StyleGAN structure}
\vspace{-3mm}
\end{figure}

The structure of QC-StyleGAN is illustrated in Fig. \ref{fig:structure}. As mentioned, it consists of two sub-networks: the mapping network $\mathcal{M}_0$ borrowed from the standard StyleGAN and a synthesis one $\mathcal{G}$. We will skip $\mathcal{M}_0$ and focus on the design of $\mathcal{G}$. We build $\mathcal{G}$ from the standard synthesis network $\mathcal{G}_0$ with minimal modifications to keep similar high-quality outputs when $q = 0$.
It consists of $N$ synthesis blocks, corresponding to different resolutions from coarse to fine. Since image degradations affect high-level details, we revise only the last $M$ layers of $\mathcal{G}$ to input the quality code $q$ and synthesize the corresponding degradation. We empirically found $M = 2$ enough to cover all common degradations. 

For each revising synthesis block, we pick the output of the last convolution layer, often named $conv1$, to revise by adding feature residuals conditioned on the quality input $q$. To do so, we introduce a novel network block, called DegradBlock, and plug it in as illustrated in Fig. \ref{fig:structure}. Let us denote the input of $conv1$ as a feature map $f \in \mathbb{R}^{C \times H \times W}$ with $C$ as the number of channels and $(H,W)$ as the spatial resolution. DegradBlock, denoted as a function $DB(\cdot)$, inputs $f$ and $q \in \mathbb{R}^{D_q}$ and outputs a residual $r \in \mathbb{R}^{C' \times H \times W}$ with $C'$ as the number of output channels of $conv1$. 

When $q = 0$, we expect $\mathcal{G}$ to behave similar to $\mathcal{G}_0$. Hence, in that case, we enforce the network features unchanged, or there is no residual:
\begin{equation}
DB(f, 0) = 0 \quad \forall f.
\label{eq:DB0}
\end{equation}
Also, when magnifying $q$, we expect the strength of image degradation, implied by the residual $r$, to increase. From those desired properties, we propose the structure of DegradBlock as in Fig. \ref{fig:degradblock}. It comprises the following components: 
\begin{enumerate}
\item A convolution layer $c$ with stride 1 to change the number of channels to be a multiple of $D_q$, denoted as $L * D_q$:
    \begin{equation}
         r_1 = c(f) \in \mathbb{R}^{LD_q \times H \times W}.
    \end{equation}
\item A linear combinator $\phi$ that splits the previous output into $D_q$ $L$-channel tensors $\{r_1^{(i)}\}$, then computes their linear combination using the weights defined by $q$:
    \begin{equation}
         r_2 = \phi(r_1) = \sum_{i=1}^{D_q} q_i * r_1^{(i)} \in \mathbb{R}^{L \times H \times W}.
    \label{eq:linearcomb}
    \end{equation}
\item A projection module $\pi$ to refine $r_2$ and change its number of channels to $C'$:
    \begin{equation}
         r = \pi(r_2) \in \mathbb{R}^{C' \times H \times W}.
    \end{equation}
    We design $\pi$ as as a stack of $P$ convolution layers with stride 1. To increase non-linearity, we put a ReLU activation after each convolution layer, except the last one. Note that when $q = 0$, we expect $r = 0$ (Equation \ref{eq:DB0}) and also have $r_2 = 0$ (according to Equation \ref{eq:linearcomb}). It leads to $\pi(0) = 0$. We ensure it by simply setting the convolution layers to have no bias.
\end{enumerate}

This design is inspired by PCA, unlike the common-used AdaIN blocks. We first use $c$ to predict $D_q$ principal components $r_1^{(i)}$, then compute their linear combination with $q$ as the component weights.
This structure is simple but satisfies the mentioned properties. When $q = 0$, the residual is guaranteed to be 0. When magnifying $q$, the residual increases accordingly (see the Appendix):
\begin{equation}
DB(f, k*q) = k * DB(f, q) \quad \forall k \in \mathbb{R}.
\label{eq:scale}
\end{equation}

\subsection{Network training}\label{sec:loss}
Next, we discuss how to train our QC-StyleGAN. As mentioned, it differs from the standard StyleGAN only on the last two blocks of the synthesis network. Hence, we initiate our network from the pretrained StyleGAN weights and finetune only those two synthesis blocks. 

Our QC-StyleGAN is trained in two modes corresponding to sharp ($q = 0$) and degraded ($q \neq 0$) image generation. We have a sharp-image discriminator $\mathcal{D}_s$ and a degraded-image discriminator $\mathcal{D}_d$ used in each mode. In the degraded image generation mode, we augment the sharp images by a random combination of primitive degradation functions (noise, blur, image compression) to get ``real'' low-quality images for training the discriminator.

For each mode, we train the networks with similar losses as in the standard StyleGAN training. 
However, in the sharp image generation mode, we employ the standard StyleGAN $\mathcal{F}_0$ as the teacher model and apply knowledge distillation to further ensure similar sharp image outputs. We transfer knowledge in the \textit{feature} spaces instead of the output space for more efficient distillation. 
Let $S_{KD}$ denotes the set of $conv1$ layers with added DegradBlocks, $X^{(l)}_0$ and $X^{(l)}$ denote features of layer $l^{th}$ in $S_{KD}$ from the teacher and student networks. We define the extra distillation loss $\mathcal{L}_{dist}$ as follows:
\begin{equation}
\mathcal{L}_{dist} = \sum _{l\in S_{KD}} \|X^{(l)}_0 - X^{(l)}\|^2.
\end{equation}
This distillation loss is added to the final loss with a weighting-hyper-parameter $\lambda_{KD}$.


\subsection{Inversion process}\label{sec:inversion}
After getting the QC-StyleGAN model, we next discuss how to fit any input image to its space. This process, called GAN Inversion, is a critical step in many applications such as image editing. 
While the general objective is to reproduce the input image, different methods optimize different components of the image generation process.
We follow the state-of-the-art technique named PTI \cite{roich2021pivotal} to optimize the $w$ embedding and the generator $\mathcal{G}$ to acquire both precise reconstruction and high editability. Furthermore, we also need to optimize the newly proposed quality-control input $q$. Let us revise the denotation of the synthesis network $\mathcal{G}$ as $\mathcal{G}_\theta$ with $\theta$ as its weights. Our inversion task $\mathcal{I}_\mathcal{G}$ estimates both the inputs $(w, q)$ and lightly tunes $\theta$ so that the reconstructed image is close to the input:
\begin{equation}
\mathcal{I}_\mathcal{G}(I, \theta_0) = (w^*, q^*, \theta^*) = \argmin_{w,q,\theta} d(\mathcal{G}_\theta(w,q), I) \qquad \qquad \text{given that} \: \|\theta - \theta_0\| < \epsilon,
\end{equation}
where $I$ is the input image, $d(\cdot)$ is a distance function, $\theta_0$ is the network weights acquired from Section \ref{sec:loss}, and $\epsilon$ is some threshold restricting the network weight change.

PTI \cite{roich2021pivotal} proposes a two-step inversion process. It first optimizes the embedding $w$ using the initial model weights $\theta_0$ (stage-1), then keeps the optimized embedding and finetunes $\theta$ (stage-2). We can adapt that process to QC-StyleGAN, with a small change to include $q$ in optimization alongside $w$. 

However, one extra requirement for this inversion, specific to QC-StyleGAN, is to have the sharp version of the reconstructed image, i.e., $\mathcal{G}_\theta(w^*, 0)$, to be high-quality. We empirically found that the naive optimization processes in PTI fail to achieve that goal. One degraded image may correspond to different sharp images, e.g., in case of motion or low-resolution blur. PTI, while manages to nicely fit the degraded input, often picks non-optimal embeddings that produce distorted corresponding sharp images. Hence, we replace its stage-1 with a training-based approach, following pSp \cite{richardson2021encoding}, with extra supervision on the sharp image domain. In this \textbf{revised stage-1}, we train two encoders to regress the embedding $w$ and the quality-code $q$ separately. Also, we load both the degraded and the corresponding sharp images for training and apply reconstruction losses on both image versions.

\subsection{Applications}\label{sec:applications}
\minisection{Image editing} \label{sec:image_dediting} The most intriguing application of StyleGAN models is to manipulate real-world images with realistic attribute changes. They can do that by applying learned editing directions in some embedding space, e.g., the $\mathcal{W}$ space. However, these standard models can only do the editing in sharp image domains. 
Our QC-StyleGAN inherits the editing ability of StyleGAN by using the same network weights except for the last $M$ synthesis blocks, which mainly affect the fine output details. However, QC-StyleGAN covers both low- and high-quality images, broadening the application domains. It can directly apply the learned editing directions of StyleGAN on low-quality images and keep their degradations unchanged. Specifically, given an input $I$ and a target editing direction $\Delta w$ learned from the standard StyleGAN in the $\mathcal{W}$ space, we can first invert the image $(w, q, \theta) = \mathcal{I}_{\mathcal{G}}(I, \theta_0)$, then generate the manipulated result $I' = \mathcal{G}_\theta(w + \Delta w, q)$. 

\minisection{Image restoration} \label{sec:restoration} This is a new functionality, which is a by-product of QC-StyleGAN design. Given a low-quality image $I$, we can fit it to QC-StyleGAN $(w, q, \theta) = \mathcal{I}_{\mathcal{G}}(I, \theta_0)$, then acquire its sharp, high-quality version by clearing the quality code: $I' = \mathcal{G}_\theta(w, 0)$.

\minisection{Degradation synthesis} QC-StyleGAN defines the degradation on the output image by a quality code $q$. It allows to revise the image degradation in various ways, such as (1) sampling a novel degradation, (2) transferring from another image, and (3) interpolating a new degradation from two reference ones.

\section{Experiments}\label{sec:exp}
\subsection{Experimental setup}\label{sec:exp_setup}
\minisection{Datasets} We conduct experiments on the common datasets used by StyleGAN, including FFHQ, AFHQ-Cat, and LSUN-Church. FFHQ \cite{karras2019stylegan} is a large dataset of 70k high-quality facial images collected from Flickr, introduced since the first StyleGAN paper. 
We will use the FFHQ images with the resolution $256\times256$. AFHQ \cite{choi2020starganv2} is a HQ dataset for animal faces with image resolution $512\times512$. 
We demonstrate our method using its Cat subset with about $5000$ images. Finally, LSUN-Church is a subset of the LSUN \cite{yu2015lsun} collection. It has about 126k images of complex natural scenes of church buildings at the resolution $256\times256$. We will use LSUN-Church only for image generation since its inversion results even on sharp image domain are not satisfactory \cite{dinh2021hyperinverter}.

\minisection{Synthesis network} We use StyleGAN2-Ada as reference to implement our QC-StyleGAN. The quality code has size $D_q = 16$. In DegradBlock, we use $L = 32$ and $P = 3$. The weight for the distillation loss $\lambda_{KD} = 3$. Our networks were trained using the same settings as in the original work until converged. Details of this training process will be provided in the Appendix.


\subsection{Image generation}
We compare the quality of our QC-StyleGAN models with their StyleGAN2-Ada counterparts in Table \ref{tab:fid}, using the FID metric. As can be seen, our models have equivalent results to the baselines when generating sharp images. However, while the common StyleGAN2-Ada models cannot produce degraded images, ours can generate such images directly with good FID scores. Even when training new StyleGAN2-Ada models dedicated on degraded images, their FID scores are worse than ours.

\begin{figure}[t]
    \centering
    \includegraphics[width=.97\textwidth]{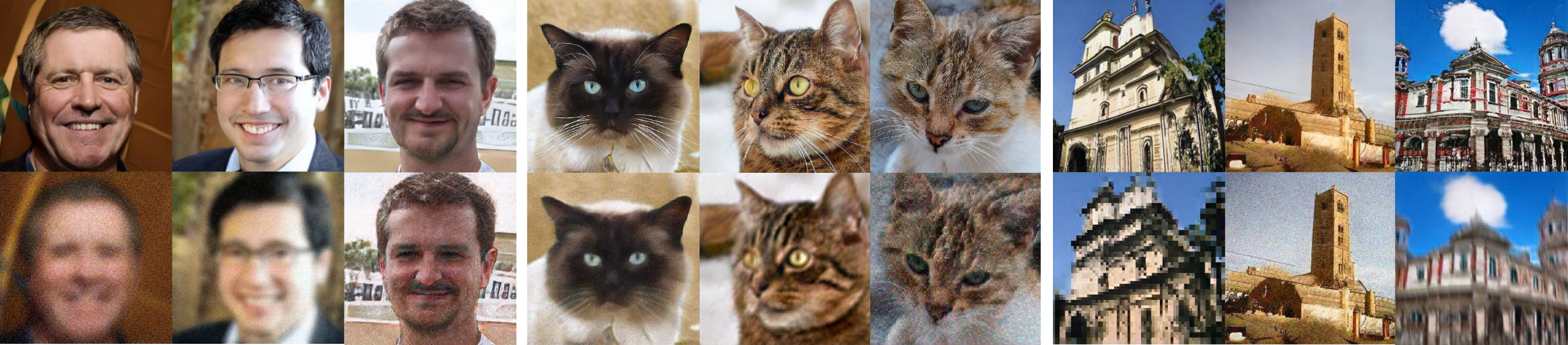}
    \vspace{-1mm}
    \caption{Sample images generated by our models on FFHQ (left), AFHQ-Cat (middle), and LSUN-Church (right). For each sample, we provide a pair of sharp (top) and degraded (bottom) images.} 
    \label{fig:im_gen}
    \vspace{-4mm}
\end{figure}

Fig. \ref{fig:im_gen} provides some samples synthesized by our networks. For each data sample, we synthesize a degraded image and the corresponding sharp version. As can be seen, the sharp images look realistic, matching the standard StyleGAN's quality. The degraded images match the sharp ones in content, and the degradations are diverse, covering noise, blur, compression artifacts, and their mixtures.

\setlength{\tabcolsep}{9pt}
\begin{table}[t]
    \centering
    \caption{\textbf{FID scores} of our QC-StyleGAN models, in comparison with the baseline StyleGAN2-Ada (SG2-Ada) \cite{karras2020ada}, on sharp and degraded image generation modes. `*' means a new, separate SG2-Ada model trained on degraded images.}
\begin{tabular}{ccccccc}
\toprule
\multirow{2.5}{*}{\textbf{FID}} & \multicolumn{2}{c}{\textbf{FFHQ (256$\times$256)}} & \multicolumn{2}{c}{\textbf{AFHQ Cat (512$\times$512)}} & \multicolumn{2}{c}{\textbf{LSUN Church (256$\times$256)}} \\
\cmidrule(lr){2-3}\cmidrule(lr){4-5}\cmidrule(lr){6-7}
& SG2-Ada & Ours & SG2-Ada & Ours & SG2-Ada & Ours\\
\midrule
Sharp & 3.48 & 3.65 & 3.55 & 3.56 & 3.86 & 3.61\\
Degraded & 4.38* & 3.23 & 4.70* & 3.91 & 5.16* & 4.58 \\
\bottomrule
\end{tabular}
\label{tab:fid}
\vspace{-2mm}
\end{table}

\subsection{GAN inversion and Image editing}\label{sec:exp_edit}
We now turn to evaluate the effectiveness of our proposed GAN inversion technique (Section \ref{sec:inversion}) and image editing (Section \ref{sec:applications}) on low-quality image inputs. 

\begin{figure}[t]
    \centering
    \includegraphics[width=.95\textwidth]{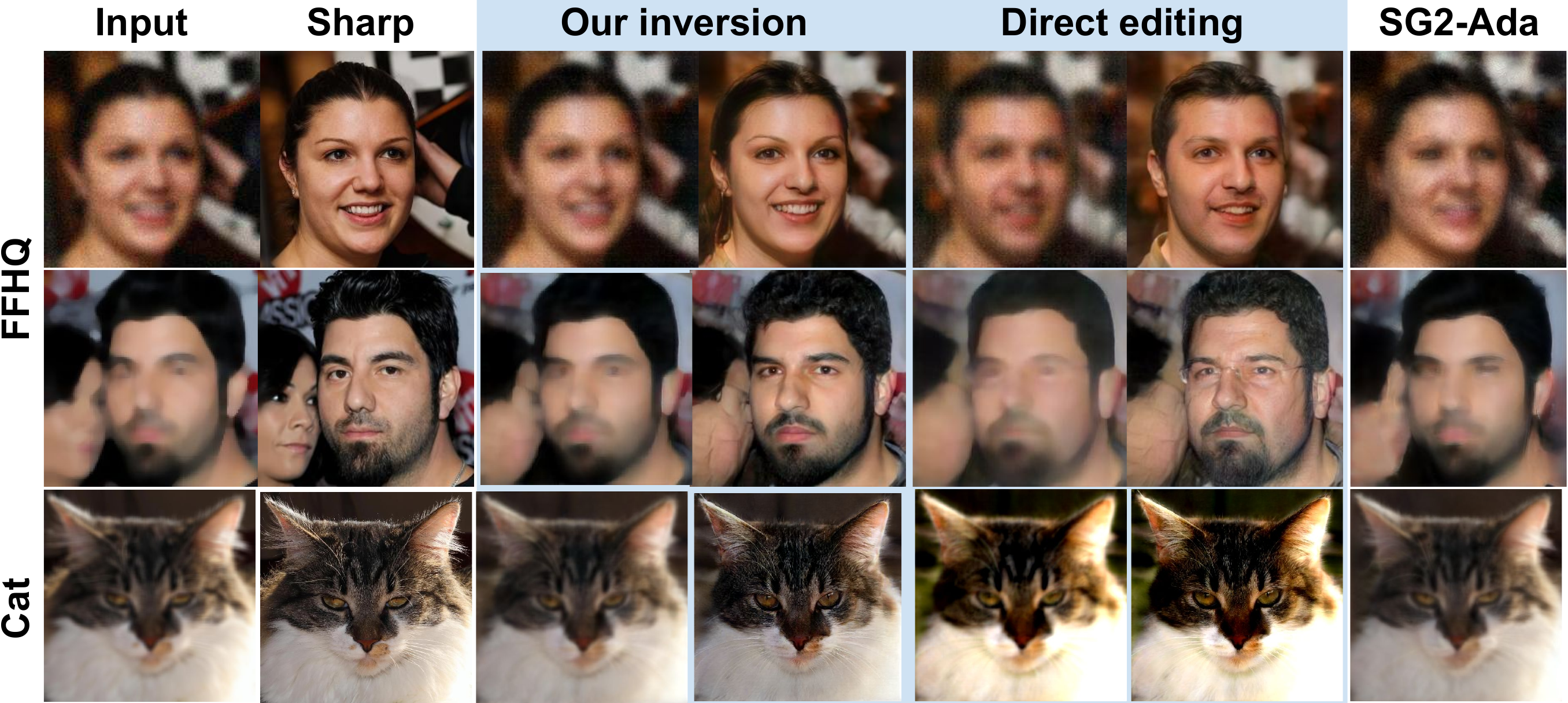} 
    \caption{\textbf{GAN Inversion and Image editing.} From a degraded input image ($1^{st}$ col.), we can fit it into our model to get a similar reconstructed degraded image ($3^{rd}$ col.), which is more accurate than from StyleGAN2-Ada (last col.). The corresponding sharp image ($4^{th}$ col.) is close to the real one ($2^{nd}$ col.). We can apply image editing directly on the degraded image ($5^{th}$ col.) and its sharp version ($6^{th}$ col.) matches the change. From top to bottom we apply gender, age, and color change.} 
    \label{fig:editing}
    \vspace{-2mm}
\end{figure}

With the model trained on the FFHQ dataset, we use the CelebA-HQ~\cite{liu2015deep, karras2018progressive} test set for evaluation. With the models trained on AFHQ-Cat, we employ its corresponding test set for testing. For each test set, we apply different image degradations to the original images to obtain the low-quality images. Our QC-StyleGAN model can fit well to such degraded inputs with the average PSNR of reconstructed images as 29.47dB and 28.91dB for CelebA-HQ and AFHQ Cat, respectively. We also try to apply the image editing directions learned for StyleGAN2-Ada by InterfaceGAN \cite{shen2020interfacegan} (face) and SeFa \cite{shen2021closedform} (cat) to manipulate the degraded images with QC-StyleGAN. The qualitative results in Fig. \ref{fig:editing} confirm the effectiveness of such a direct image editing scheme. Note that while we can also do GAN inversion with the StyleGAN2-Ada model on these inputs, its inversions cannot be converted to sharp. Also, direct manipulation on StyleGAN2-Ada's inversions may introduce unrealistic artifacts (see the Appendix), making StyleGAN inferior to QC-StyleGAN in handling low-quality inputs.  


\subsection{Image restoration}

As mentioned in \Sref{sec:restoration}, a nice by-product of our QC-StyleGAN is a simple but effective image restoration technique. We examine it on the degraded CelebA-HQ images on five common restoration tracks, including deblurring, super-resolution, denoising, JPEG removal, and multiple-degraded restoration. We also compare it with the state-of-the-art image restoration methods. For image restoration networks such as NAFNet \cite{chen2022simple} and MPRNet \cite{zamir2021multi}, we re-train the models on our degraded images using their published code with default configuration. For GAN-based methods like HiFaceGAN \cite{yang2020hifacegan} and PULSE \cite{menon2020pulse}, we use their provided pre-trained models. The results are reported in \Tref{tab:quan_restoration}.

Among the common metrics for this task, we find LPIPS more reliable and close to human perception. Although our method is not tailored to handle this restoration task specifically, it performs reasonably well and outperforms many baseline methods in each task. Particularly, QC-StyleGAN provides the best LPIPS score when having multiple degradations in the input. Also, we find that one degraded image may correspond to multiple possible sharp images. Our restoration results sometimes look reasonable but do not match the ground-truth, severely hurting our LPIPS scores.

To avoid the mismatching ground-truth issue, we also use NIQE \cite{mittal2012making}, which is a no-reference image quality metric. QC-StyleGAN provides the best NIQE score in nearly all tracks. It confirms that our image restoration can produce the highest output quality in terms of naturalness \cite{mittal2012making} while still maintaining comparable perceptual similarity \cite{zhang2018unreasonable} compared to the competitors.

\begin{figure}[t]
    \centering
    \includegraphics[width=.95\textwidth]{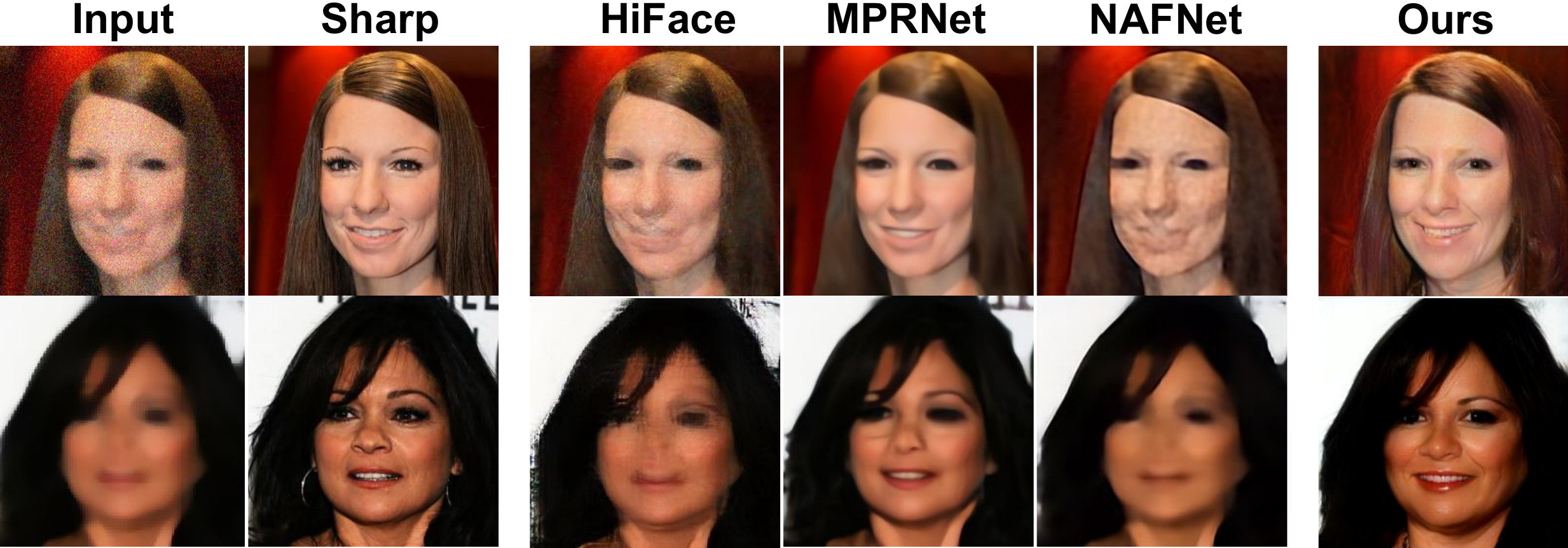} 
    \vspace{-2mm}
    \caption{Comparison between our method and image restoration baselines on CelebA-HQ dataset.} 
    \label{fig:restore}
    \vspace{-2mm}
\end{figure}

Fig. \ref{fig:restore} provides restoration results from our method and the image restoration baselines on two extremely degraded images. Our algorithm manages to return sharp and detailed images, while the others fail to handle such images and show clear artifacts on their recovered images.

\begin{table}[t]
\begin{center}
\setlength{\tabcolsep}{4pt}
\begin{tabular}{lccccc} 
\toprule
& \multicolumn{5}{c}{Track} \\
\cmidrule{2-6}
& {Blur} & {Super-res.} & {Noise} & {JPEG comp.} & {Multiple-deg.} \\
\midrule
HiFaceGAN \cite{yang2020hifacegan} & \underline{5.95} / 0.216 & \textbf{5.32} / \underline{0.125} & 6.01 / \underline{0.126} & \underline{4.95} / \textbf{0.053} & \underline{5.916}   / 0.364 \\
ESRGAN \cite{wang2021realesrgan} & - & 6.35 / 0.148 & - & - & - \\
DnCNN \cite{zhang2017beyond} & - & - & 6.93 / \textbf{0.080} & - & - \\
MPRNet \cite{zamir2021multi} & 8.12 / \textbf{0.194} & 6.73 / 0.230 & 7.33 / 0.143 & 7.64 / 0.128 & 8.97 / \underline{0.299}\\
NAFNet \cite{chen2022simple} & 7.42 / 0.599 & 6.73 / 0.586 & \underline{5.58} / 0.557  & 5.92 / 0.561 & 7.30 / 0.611\\
PULSE \cite{menon2020pulse} & - & 6.29 / 0.296 & - & - & - \\
mGANPrior \cite{gu2020image} & - & 6.02 / 0.265 & - & - & - \\
GLEAN \cite{chan2021glean} & - & 7.29 / \textbf{0.072} & - & - & - \\
Ours & \textbf{5.83} / \underline{0.195} & \underline{5.45} / 0.177 & \textbf{5.41} / 0.183 & \textbf{4.51} / \underline{0.118} & \textbf{5.64} / \textbf{0.260} \\
\bottomrule
\end{tabular}
\end{center}
\caption{NIQE \cite{mittal2012making} and LPIPS \cite{zhang2018unreasonable} scores of image restoration methods on five restoration tracks on the CelebA-HQ dataset. For both metrics, lower value means better. The \textbf{best} and \underline{runner-up} values are marked in bold and underline, respectively. The mark '-' means the method is not applicable.}
\vspace{-3mm}
\label{tab:quan_restoration}
\end{table}


\subsection{Degradation synthesis}
\begin{figure}[t]
    \centering
    \includegraphics[width=.95\textwidth]{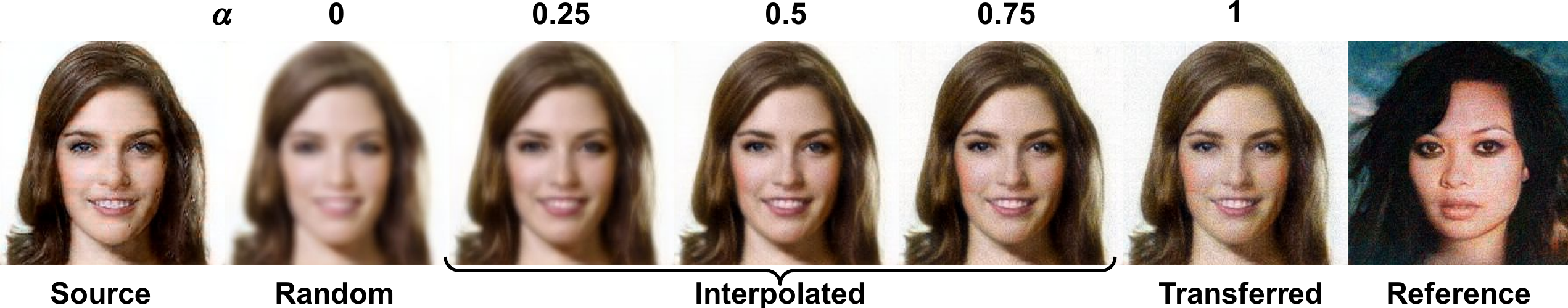} 
    \vspace{-1mm}
    \caption{Degradation synthesis} 
    \label{fig:deg_syn}
    \vspace{-5mm}
\end{figure}

We provide an example of our proposed image degradation synthesis (Section \ref{sec:applications}) in Fig. \ref{fig:deg_syn}. From a source image with JPEG compression artifacts, we change its image degradation to a novel random one (blur, $2^{nd}$ col.) or copy the degradation from a reference image (noise, $6^{th}$ col.). We can also smoothly interpolate in-between degradations, using an interpolation factor $\alpha \in [0, 1]$ ($3-5^{th}$ col.).

\subsection{Quality-code-based Degradation Classification}
To verify the quality of QC-StyleGAN in degradation estimation, we conduct experiments of training linear classifiers to predict whether an image is blurry, or at which blur level, based on its quality code $q$. For each experiment, we use 1000 facial images for training and 200 images for testing. When we train the classifier to detect if an image is blurry, the accuracy is \textbf{97.9\%}. When we divide the degree of blurring into 5 levels for the classifier to predict, the accuracy is 85\%. When we use 10 blur levels, the accuracy is still high (77\%), confirming QC-StyleGAN as a quality degradation estimator.

\subsection{Stability of Inversion and Editing}
We conduct experiments to verify the stability of our inversion and editing on degraded CelebA-HQ inputs images. We tried the editing magnitudes $\pm3$ for 3 editing tasks on gender, age, and smiling. For each input image and each editing operator, we executed the operator 3 times to get 3 manipulated outputs and compare them pairwise using the PSNR and LPIPS metrics. The manipulated degraded images are pretty similar, with the PSNR score $44.42 \pm 2.71$ and the LPIPS score $0.004 \pm 0.003$. When recovering the sharp version of these images, the PSNR score is still high ($41.84 \pm 2.07$) and the LPIPS score is still very good ($0.006 \pm 0.005$).

We generated some quantitative videos and provided them in  \href{https://bit.ly/3JEIz3C}{this link}. For each video, we show the same manipulation in 6 runs. As can be seen, our manipulated results are quite stable, with minor flickers mainly appearing on the background or the hair region. If we mask out the background and keep only the face region (the hair region is still kept), the scores in the previous experiments get improved: For degraded images, the PSNR score is $47.18 \pm 2.27$, and the LPIPS score is $0.0013 \pm 0.0006$. For sharp-recovered images, the PSNR score is $43.67 \pm 2.28$, and the LPIPS score is $0.0028 \pm 0.0013$. It confirms that our inversion and editing are pretty stable on the target object.



\subsection{Ablation studies}
In this section, we investigate the effectiveness of our model design on the FFHQ dataset. First, we try other ways to inject $q$ into the network, e.g., concatenating $q$ with the latent input $z$, the embedding $w$, or via DualStyleGAN structure \cite{yang2022Pastiche}, but find clear mismatches in image content of generated sharp and degraded images of the same code $w$. Second, we try our network training without the distillation loss, and the FID-sharp is very high at $11.32$. Third, we try a AdaIN-style design for the DegradBlocks, and its FID-sharp is $4.41$. If we apply DegradBlock at the last synthesis block ($M = 1$), the FID-sharp is $6.38$. If we reduce $D_q$ to $8$, that FID-sharp increases ($4.54$) even if we expand $L$ to $64$ for a similar computation cost. Extra discussions will be included in the Appendix.

\subsection{Inference time}
In this section, we report the running time of the proposed models in \Tref{tab:runtime}. For the restoration tasks, we report the running time of the two components of the method, including pSp (stage-1) and PTI optimization (stage-2).
\begin{table}[ht]
    \setlength{\tabcolsep}{8pt}
    \caption{\textbf{Average running times} of our methods on two tasks, including image generation and image inversion (pSp and PTI optimization). Note that the performance of pSp is the same on both resolutions because it resizes the input to $256 \times 256$ before running.}
    \centering
    \begin{tabular}{lcc}
        \toprule
         \multicolumn{1}{c}{\multirow{2}{*}{Model}} & \multicolumn{2}{c}{Time (s)}\\
         \cmidrule(lr){2-3}
              & 256 $\times$ 256 & 512 $\times$ 512  \\
         \midrule
         QC-StyleGAN & 0.08 $\pm$ 0.01 & 0.20 $\pm$ 0.04\\
         pSp & 0.13 $\pm$ 0.01 & 0.11 $\pm$ 0.01 \\
         PTI-opt & 83.40 $\pm$ 2.68 & 222.29 $\pm$ 23.24\\
        \bottomrule     
    \end{tabular}
    \label{tab:runtime}
\end{table}

\section{Conclusions and future work}
This paper presents QC-StyleGAN, a novel image generation structure with quality-controlled output. It inherits the capabilities of the standard StyleGAN but extends to cover both high- and low-quality image domains. QC-StyleGAN allows direct manipulation on in-the-wild, low-quality inputs without quality changes. It offers novel functionalities, including image restoration and degradation synthesis.

\textbf{Limitations.} Although we employed many image degradations in training QC-StyleGAN, they might not cover all in-the-wild degradations. Also, we only implemented QC-StyleGAN from StyleGAN2-Ada. We plan to improve on these aspects in the future.


\textbf{Potential negative societal impacts.} Our work may have negative societal impacts by allowing better image manipulation on in-the-wild data. However, image forensics is an active research domain to neutralize that threat. We believe our work's benefits dominate its potential risks.


{
\small
\bibliographystyle{unsrt}
\bibliography{qcstylegan}
}



\appendix

\section{Proofs}
\subsection{Proof of Equation ~\ref{eq:scale}} 

First, let us consider the linear combinator $\phi$. When we scale the input $q$, its value is scaled accordingly:
\begin{equation}\label{eq:scale_phi}
\phi(r_1, k * q) = \sum_{i=0}^{D_q} (k * q_i) * r_1^{(i)} = k * \sum_{i=0}^{D_q} q_i * r_1^{(i)} = k * \phi(r_1, q) \quad \forall k \in \mathbb{R}.
\end{equation}

Next, the projector $\pi$ is a composition of $P' = 2P-1$ layers:
\begin{equation}
\pi = \pi^{(P')} \circ \pi^{(P'-1)} \circ ... \circ\pi^{(1)}
\end{equation}
where each layer $\pi^{(i)}$ is either (1) a convolution layer with stride 1 and no bias $\pi_{conv}^{(i)}(x) = W^{(i)} * x$, with $W^{(i)}$ is a weighting matrix, or (2) a RELU layer $\pi_{RELU}^{(i)}(x) = max(x, 0)$. Both of layer types are homogeneous functions with degree 1 since:
\begin{equation}
\begin{aligned}
\pi_{conv}^{(i)}(k*x) & = W^{(i)} * (k * x) = k * (W^{(i)} * x) = k * \pi_{conv}^{(i)}(k) \quad \forall k \in \mathbb{R},\\
\pi_{RELU}^{(i)}(k*x) & = max(k*x, 0) = k * max(x, 0) = k * \pi_{RELU}^{(i)}(x) \quad \forall k \in \mathbb{R}.
\end{aligned}
\end{equation}
Hence:
\begin{equation}\label{eq:scale_pi}
\begin{split}
\pi(k*x) & = \pi^{(P')}(\pi^{(P'-1)}(...(\pi^{(1)}(k*x)))) \\
         & = \pi^{(P')}(\pi^{(P'-1)}(...(k * \pi^{(1)}(x)))) \\
         & = ... \\
         & = \pi^{(P')}(k*\pi^{(P'-1)}(...(\pi^{(1)}(x)))) \\
         & = k * \pi^{(P)}(\pi^{(P'-1)}(...(\pi^{(1)}(x)))) \\
         & = k * \pi(x) \qquad \qquad \qquad  \qquad \qquad \forall k \in \mathbb{R}.
\end{split}
\end{equation}
That means the projector $\pi$ is also a homogeneous function with degree 1.

Finally, we consider the entire DegradBlock:
\begin{align*}
DB(f, k * q) = \pi(\phi(r_1, k * q)) & = \pi(k * \phi(r_1, q)) \tag{from Equation \ref{eq:scale_phi}} \\
              & = k * \pi(\phi(r_1, q)) \tag{from Equation \ref{eq:scale_pi}} \\
              & = k * DB(f, q) \quad \forall k \in \mathbb{R}.
\end{align*}
Thus, the lemma in Equation \ref{eq:scale} holds.

\section{Training details}
\paragraph{Hyperparameters}
We build upon the official Pytorch implementation of StyleGAN2-Ada by Karras et al., from which we inherit most of the training details, including weight demodulation, path length regularization, lazy regularization, style mixing regularization, equalized learning rate for all trainable parameters,  the exponential moving average of generator weights, non-saturating logistic loss with $R_1$ regularization, and more. We use Adam optimizer with $\beta_1$ = 0, $\beta_2$ = 0.99, and $\epsilon$ = $10^{-8}$. The quality code has size $D_q = 16$. In DegradBlock, we use $L = 32$ and $P = 3$. The weight for the distillation loss $\lambda_{KD} = 3$. We also report other details for each training in Table \ref{tab:hyperparam}.

For the pSp model, we use the github project with some modifications to change the generative model to our QC-StyleGAN and add an additional encoder for quality code inversion. Besides, all the network architecture and hyperparameters are kept intact.

\paragraph{Training environment}
We ran our QC-StyleGAN training on a DGX SuperPOD node with 8 Tesla A100 GPUs, using Pytorch 1.7.1 (for comparison methods), CUDA 11.1, and cuDNN 8.0.5. We use the official pre-trained Inception network to compute FID. 

For the image restoration task, we train pSp and run PTI on a single Nvidia V100. It took about 3 days for the pSp training to converge.

\paragraph{Datasets}
We train our QC-StyleGAN on common datasets, including FFHQ, LSUN-Church, and AFHQ-Cat. For the image restoration task, we train pSp on FFHQ and AFHQ-cat datasets and test the trained models on CelebA and AFHQ-Cat validation sets, respectively. Details of the used datasets are listed in \Tref{tab:dataset_details}.

\begin{table}[ht]
\centering
\caption{Hyperparameters used in each model training.} 
\begin{tabular}{lccc}
\toprule
\textbf{Parameter}  & \textbf{FFHQ} & \textbf{AFHQ-Cat} & \textbf{LSUN-Church}  \\
\midrule
Resolution          & 256$\times$256       & 512$\times$512           & 256$\times$256               \\
Number of GPUs      & 8             & 8                 & 8                     \\
Training length     & 5M            & 5M                & 5M                
\\
Minibatch size      & 64            & 64                & 64                    \\
Minibatch stddev    & 8             & 8                 & 8                     \\
Feature maps        & $\frac{1}{2}\times$          & 1$\times$                & 1$\times$                 \\
Learning rate $\eta \times 10^3$ & 2.5           & 2.5               & 2.5                   \\
$R_1$ regularization $\gamma$ & 1             & 0.5               & 100                   \\
Mixed-precision     & \checkmark          & \checkmark              & \checkmark                  \\
Mapping net depth   & 8             & 8                 & 8                     \\
Style mixing reg.   & \checkmark          & \checkmark              & \checkmark                  \\
Path length reg.    & \checkmark          & \checkmark              & \checkmark                  \\
Resnet D            & \checkmark          & \checkmark              & \checkmark\\
Training time       & 1.5 days          & 3.5 days              & 1.5 days\\
\bottomrule
\end{tabular}
\label{tab:hyperparam}
\end{table}

\begin{table}[t]
    \centering
    \caption{\textbf{Dataset details.} Asterisks($^{\ast}$) indicate that the set is randomly sampled from the original dataset.}
    \begin{tabular}{cccc}
        \toprule
         Dataset & Resolution & \#training images & \#testing images \\
         \midrule
         FFHQ & 256 $\times$ 256 & 70000 & - \\
         CelebA$^{\ast}$ & 256 $\times$ 256 & - & 300 \\
         AFHQ-Cat & 512 $\times$ 512 & 5153 & 300 \\
         LSUN Church & 256 $\times$ 256 & 126227 & - \\
         \bottomrule
    \end{tabular}
    \label{tab:dataset_details}
\end{table}


\section{Additional Experimental Results}
\subsection{Comparison between Inversion Results of StyleGAN and QC-StyleGAN}
\begin{figure}[ht]
    \centering
    \includegraphics[width=\textwidth]{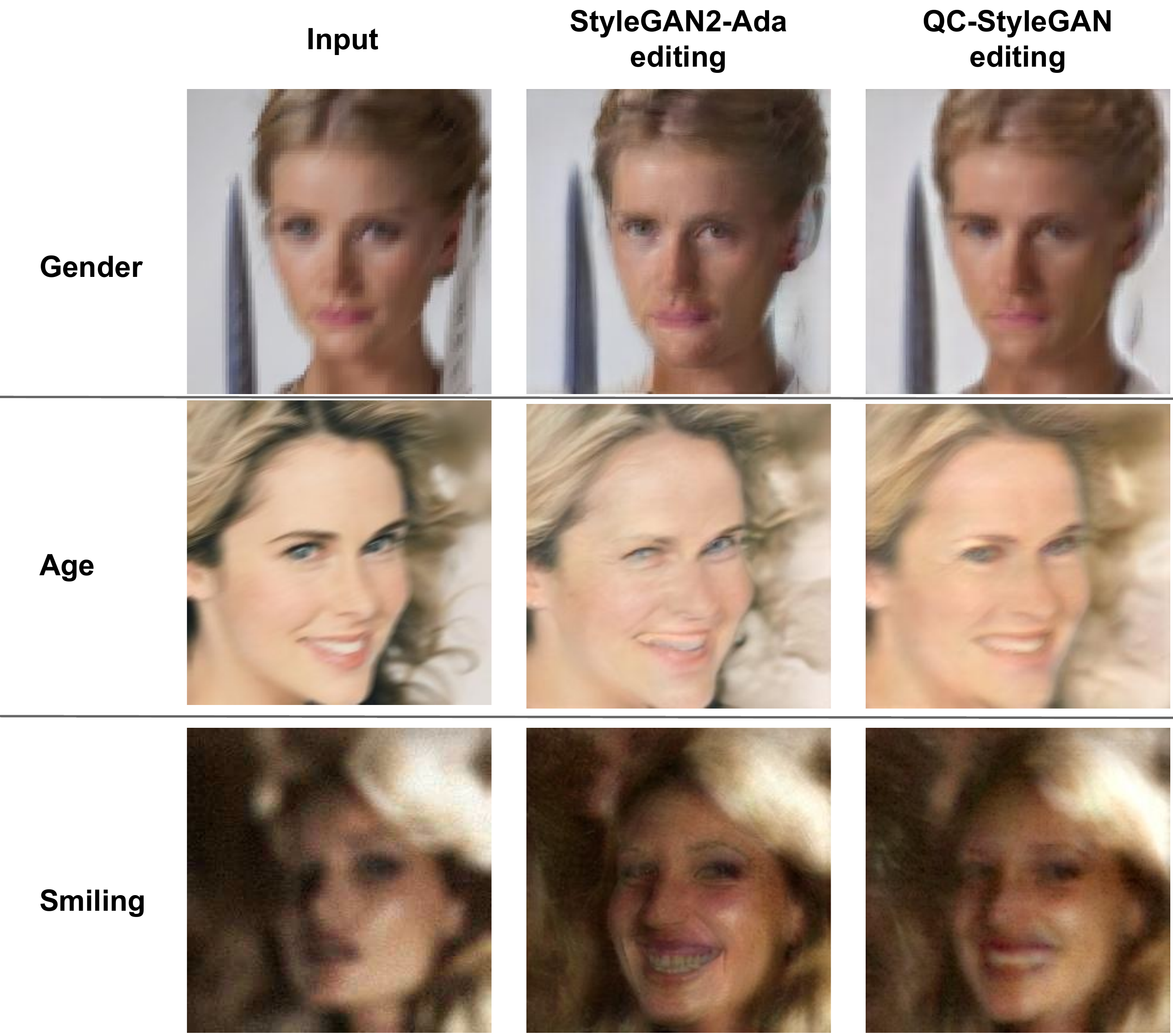}
    \caption{\textbf{Image editing comparison.} We compare image editing results based on StyleGAN2-Ada and QC-StyleGAN inversion from degraded FFHQ input images. QC-StyleGAN produces natural-looking images, while StyleGAN2-Ada sometimes produces unrealistic artifacts.}
    \label{fig:compare_edit}
\end{figure}

The differences between our inversion results and StyleGAN2-Ada inversion ones:
\begin{itemize}
    \item First, our reconstructed images can easily be converted to their sharp version. In contrast, inversion with StyleGAN2-Ada only gives us fitted degraded images.
    \item Moreover, since QC-StyleGAN models both sharp and degraded images, the inversion results often stay in-distribution, allowing good editing results. In contrast, the original StyleGAN2-Ada network only models sharp images; editing its inversion on degraded inputs might lead to unrealistic outcomes. We provide the comparison between their manipulation results in Fig. \ref{fig:compare_edit}.
\end{itemize}

\subsection{Quantitative Results on Image Editing}
We quantitatively evaluate the editability of our proposed GAN inversion method with QC-StyleGAN by measuring the amount of target attribute changes with different editing magnitudes. Given a degraded image, we synthesize its corresponding edited versions:
\begin{align}
    I_{t} = G(w + \gamma_t * d, q)
\end{align}
where $\gamma_t$ is one of the $10$ edit magnitudes uniformly picked in the range of $[-3, 3]$, $w$ and $q$ are the content and quality codes inverted by our proposed inversion method, $d$ is the editing direction explored from the latent space of QC-StyleGAN by using InterfaceGAN \cite{shen2020interfacegan}. In this test, we choose the age editing direction. To measure the age change when editing, we leverage the off-the-shelf DEX VGG \cite{rothe2015dex} model to estimate the face's age in the images. We perform this experiment on $100$ multiple-degraded images and show the average age changes for each magnitude in Table \ref{tab:editing_evaluation}. We report results for both the manipulated degraded outputs and their sharp recovered version. The age change scales consistently with the editing magnitude, confirming the editability of the latent space of our generator. Note that the age regression model was trained on sharp images; hence, it produces less significant changes on degraded images, which may not reflect the actual shift on these pictures.

\begin{table}[t]
\begin{center}
\begin{tabular}{ccc} 
\toprule
$\gamma$ & Our (degraded) & Ours (sharp) \\
\midrule
-3.0  & 4.22 & 15.04  \\
-2.33  & 3.39 & 11.79  \\
-1.67  & 2.72  & 9.16  \\
-1.0   & 1.86 & 6.56  \\
-0.33  & 1.12 &  3.5 \\
0.33   & 0.6 &  0.72 \\
1.0   & -0.25 &  -1.78 \\
1.67   & -1.19 &  -4.24 \\
2.33   & -1.97 &  -6.12 \\
3.0   & -2.5 &  -7.63 \\
\bottomrule
\end{tabular}
\end{center}
\caption{\textbf{Quantitative evaluation of editability.} We apply age editing on degraded CelebA-HQ images using different magnitudes $\gamma$ and measure the amount of age change. We test both the manipulated degraded outputs and their sharp recovered version.}
\vspace{-3mm}
\label{tab:editing_evaluation}
\end{table}



\subsection{Ablation Studies}
We report detailed results of our ablation studies in Table \ref{tab:detailed_ablation}. 

When there is no distillation loss, the training process is unstable and diverges, causing very high FID scores either when finetuning only the last $M$ synthesis blocks (\textbf{B}) or the entire network (\textbf{A}). It confirms the importance of such distillation loss in our network training. 

Next, we investigate the number of synthesis blocks to finetune $M$. When we finetune only 1 block (\textbf{C}), the FID score for sharp image generation is 6.38 and the one for degraded image generation is 7.59, which are still high. When $M = 2$ (the official implementation), the FIDs are small, and FID-sharp is close to the one from the standard StyleGAN2-Ada. Since the network performance is already satisfactory, we skip testing with $M > 2$, which requires much more computational cost.

Next, we investigate the design of DegradBlock. We find that increasing the number of layers inside the projection module does not help (\textbf{D}), and the FIDs slightly increase. 

Next, we examine the choice of the quality code size $D_q$. In our implementation, $D_q$ is set as $16$. We tried with a small quality code size with $D_q = 8$, but the FIDs are not as good, even when we increase the number of intermediate channels $L$ to 64 (\textbf{E}). It confirms that $D_q$ should be large to cover a wide range of image degradations.

Finally, we try an AdaIN-like DegradBlock (\textbf{F}) by replacing the convolution layer $c$ and the linear combinator $\phi$ with an AdaIN module. The FID scores for clean and degraded images on the FFHQ dataset are $4.41$ and $5.28$, respectively, which are significantly higher than our PCA-based DegradBlock design.

\begin{table}[ht]
\centering
\caption{Ablation studies on our QC-StyleGAN network design.}
\begin{tabular}{lrr}
\hline
\multicolumn{1}{c}{\multirow{2}{*}{\textbf{Configuration}}} & \multicolumn{2}{c}{\textbf{FID}}                         \\ \cline{2-3} 
\multicolumn{1}{c}{}                                        & \multicolumn{1}{l}{Sharp} & \multicolumn{1}{l}{Degraded} \\ \hline
\textbf{A} no distillation loss + no freezed                                                       & 11.32                     & 26.91                        \\ \hline
\textbf{B} no distillation loss + freezed                                                          & 33.45                     & 82.11                        \\ \hline
\textbf{C} $M$ = 1 layer of $G$                                                                        & 6.38                      & 7.59                         \\ \hline
\textbf{D}  DegradBlock:  stack of $P$ = 5 convolution layers                                        & 4.02                      & 4.68                         \\ \hline
\textbf{E} The quality code has size $D_q = 8$ and $L = 64$               & 4.54                      & 5.69                         \\ \hline
\textbf{F} AdaIN-style DegradBlock                                                                        & 4.41                      & 5.28                         \\ \hline
\label{tab:detailed_ablation}
\end{tabular}
\end{table}


\section{Additional Qualitative Results}
\subsection{Image generation}
\begin{figure}[ht]
    \centering
    \includegraphics[scale=0.4]{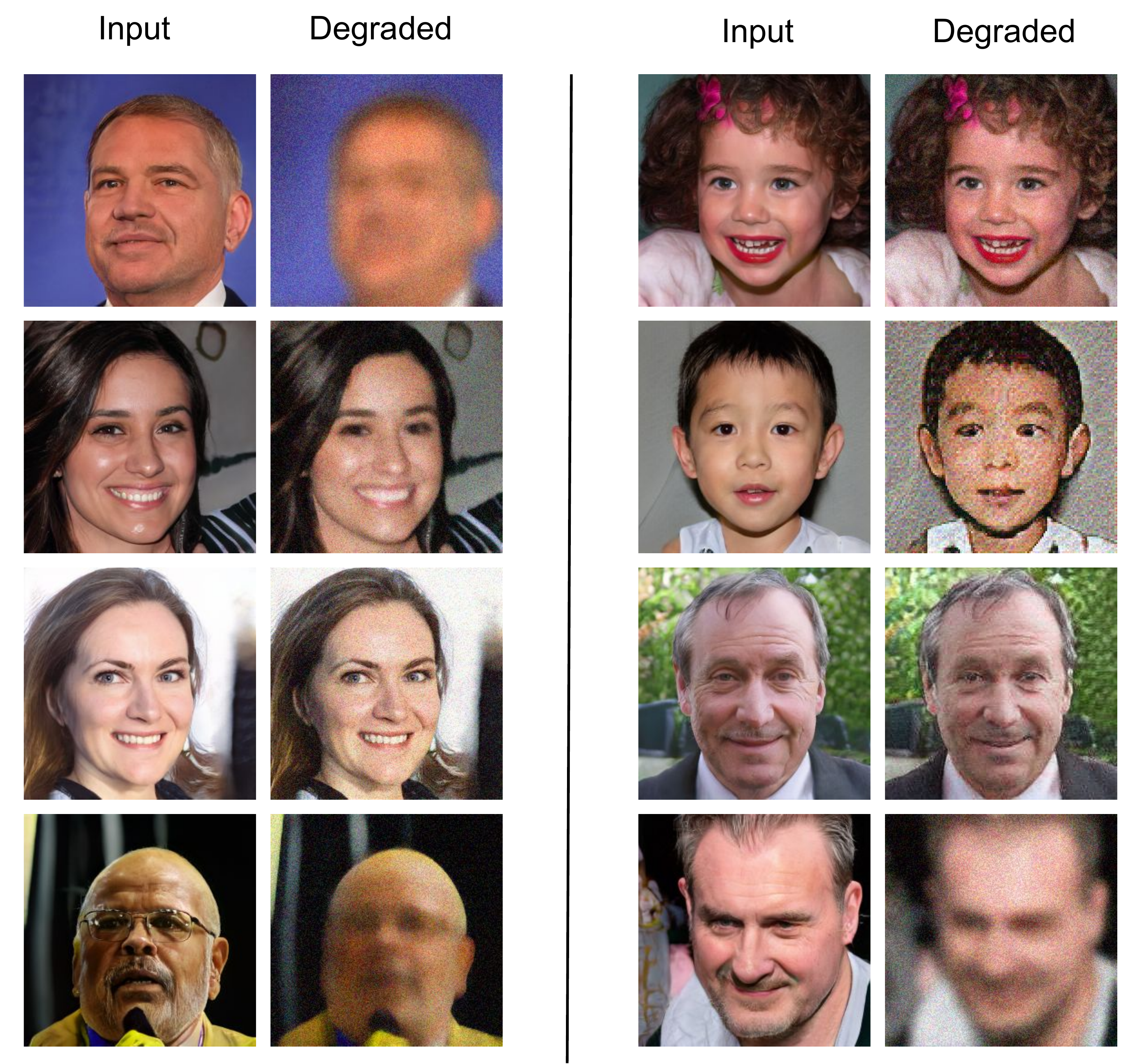}
    \caption{Sample images generated by our QC-StyleGAN trained on the FFHQ dataset.}
    \label{fig:ffhq_gen_01}
\end{figure}

\begin{figure}[ht]
    \centering
    \includegraphics[scale=0.4]{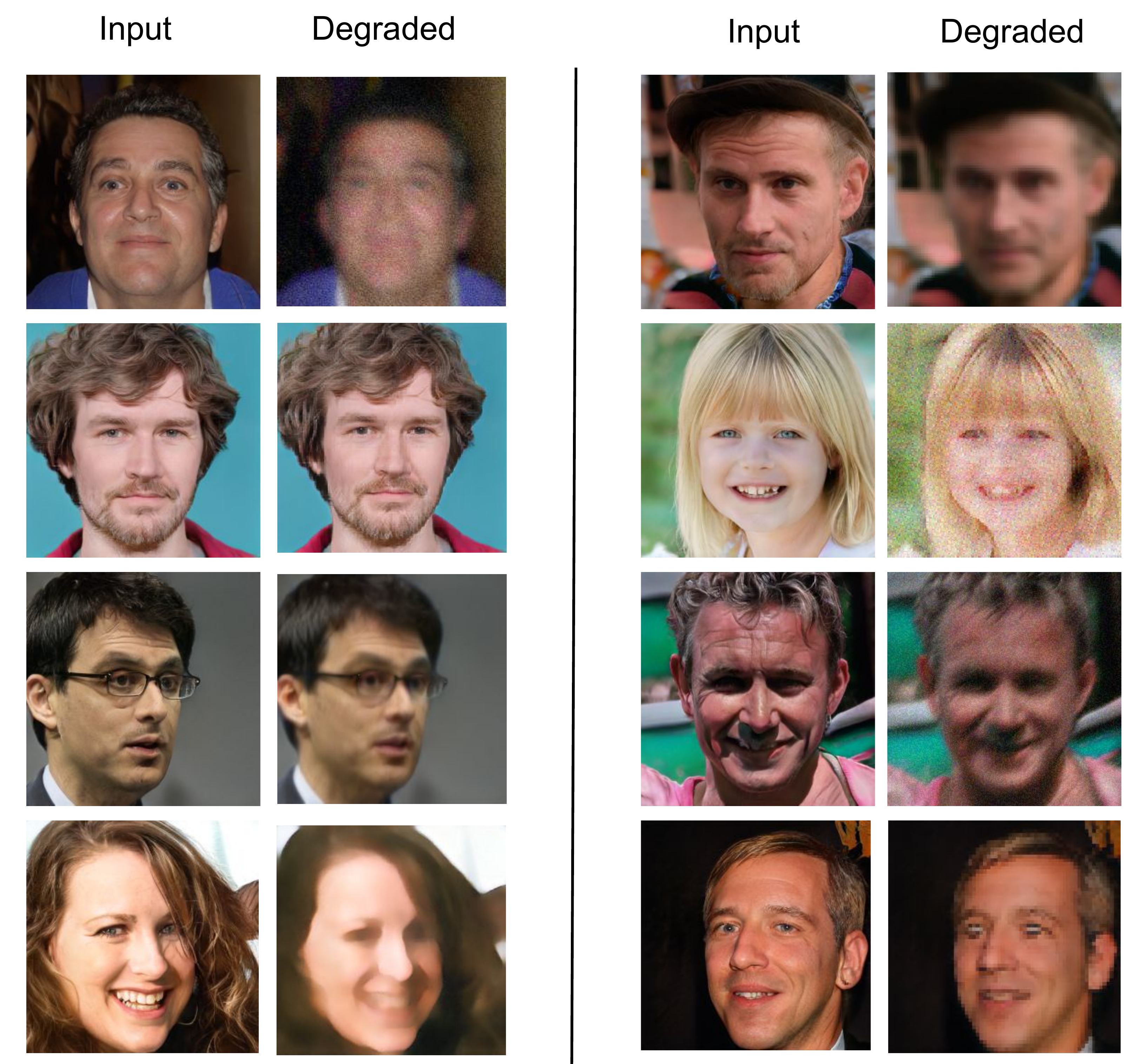}
    \caption{Sample images generated by our QC-StyleGAN trained on the FFHQ dataset.}
    \label{fig:ffhq_gen_02}
\end{figure}

\begin{figure}[ht]
    \centering
    \includegraphics[scale=0.2]{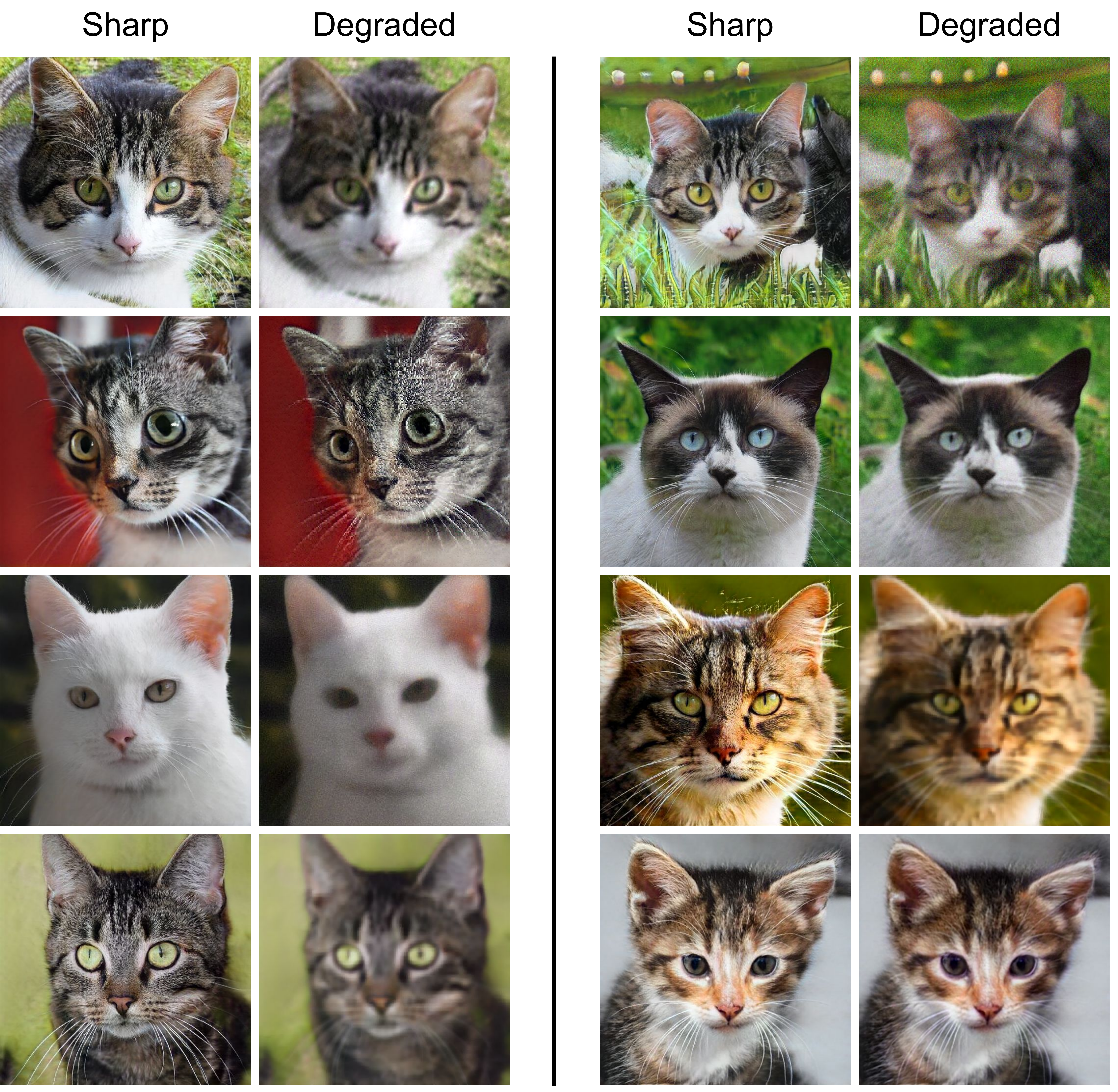}
    \caption{Sample images generated by our QC-StyleGAN trained on the AFHQ-Cat dataset.}
    \label{fig:cat_gen_01}
\end{figure}

\begin{figure}[ht]
    \centering
    \includegraphics[scale=0.2]{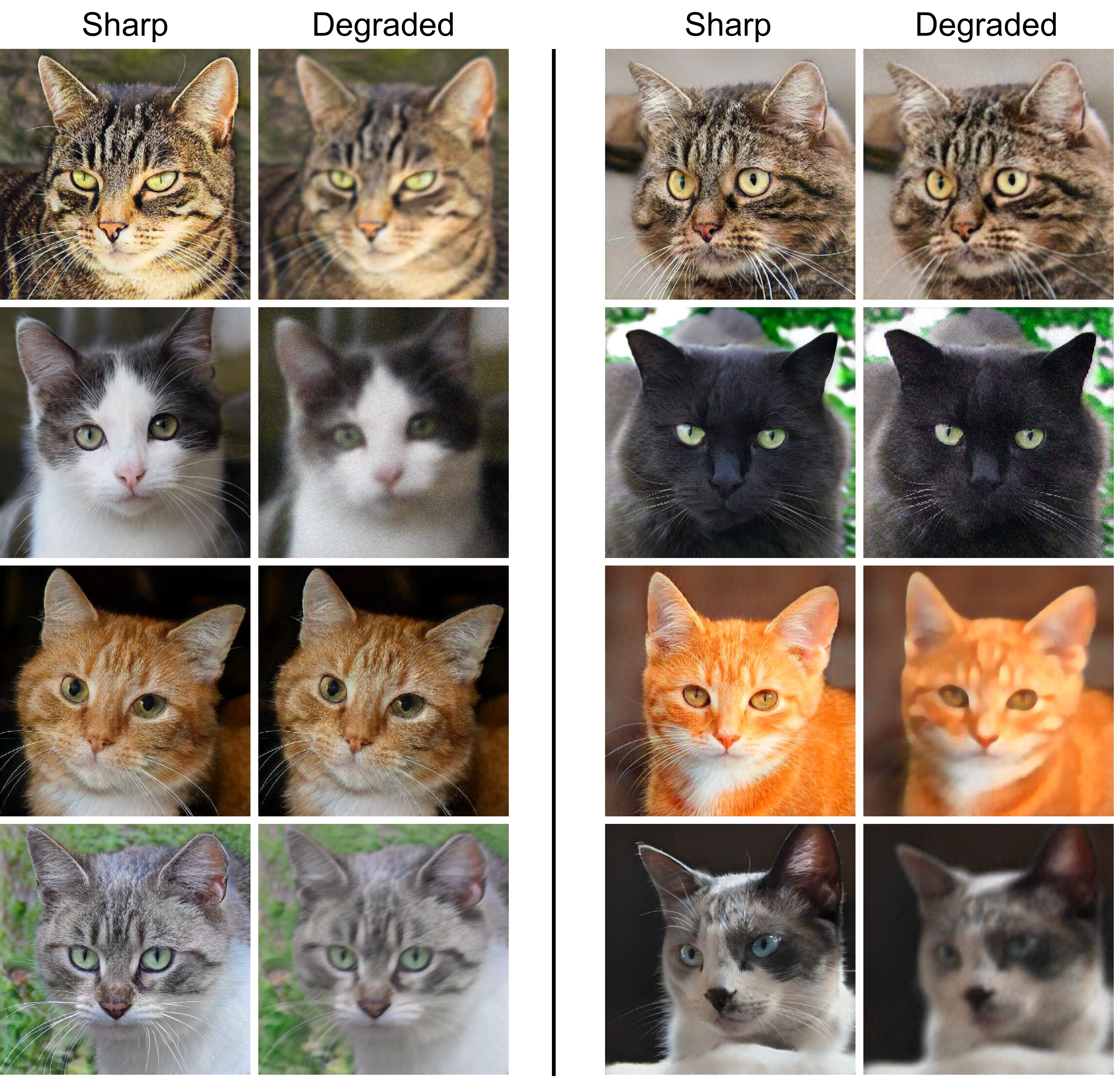}
    \caption{Sample images generated by our QC-StyleGAN trained on the FFHQ dataset.}
    \label{fig:cat_gen_02}
\end{figure}

\begin{figure}[ht]
    \centering
    \includegraphics[scale=0.4]{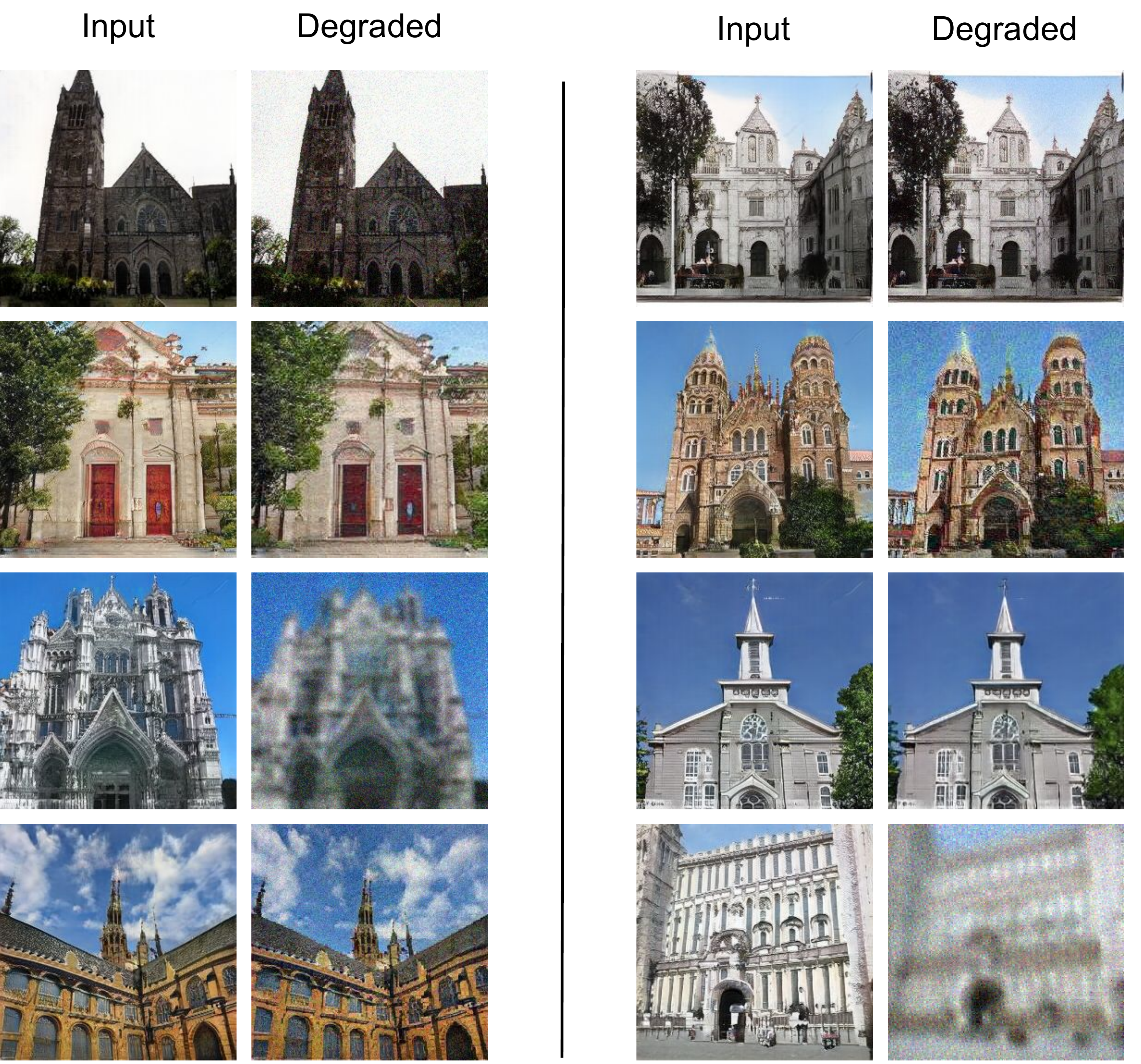}
    \caption{Sample images generated by our QC-StyleGAN trained on the LSUN-Church dataset.}
    \label{fig:church_gen_01}
\end{figure}

\begin{figure}[ht]
    \centering
    \includegraphics[scale=0.4]{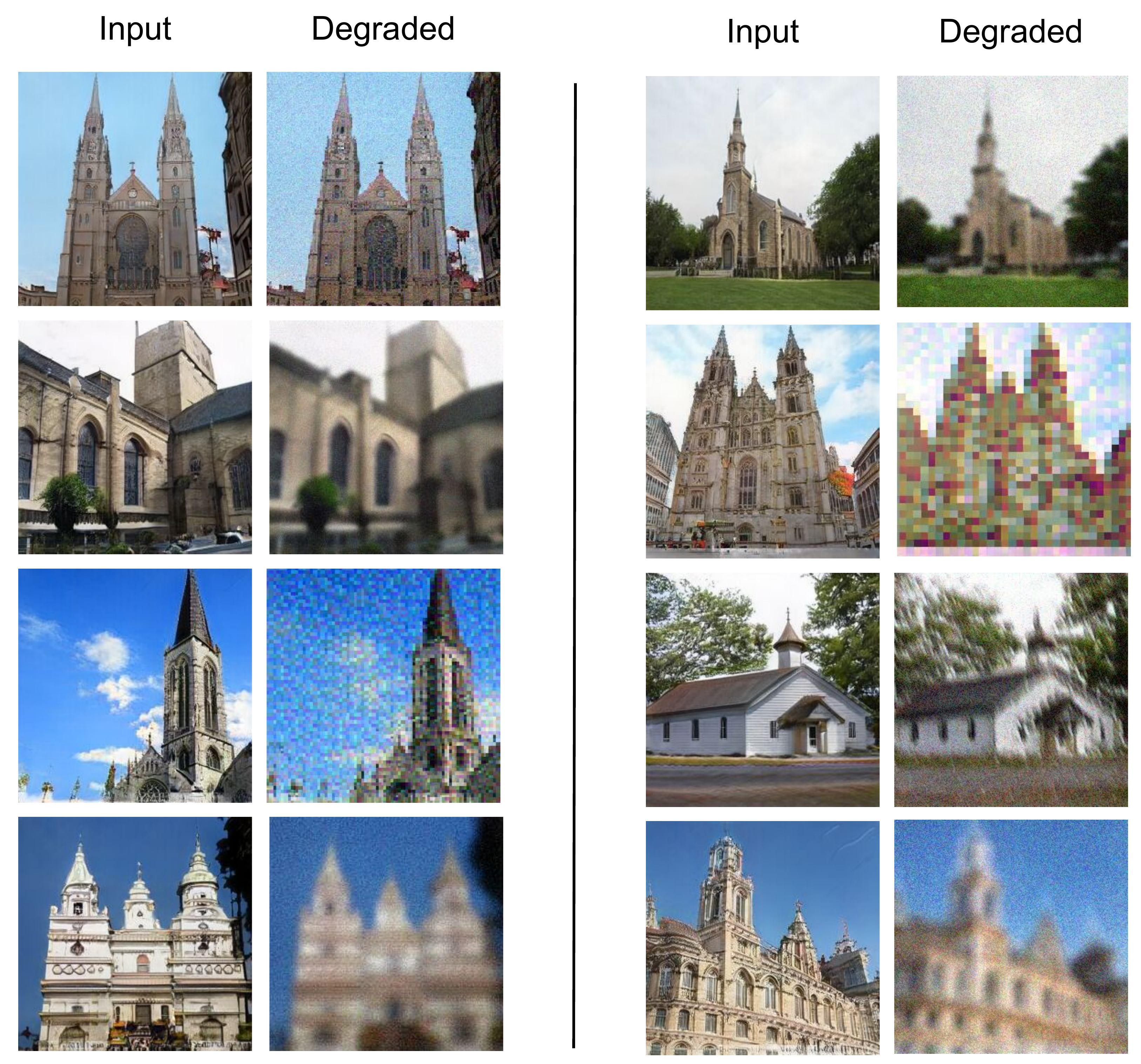}
    \caption{Sample images generated by our QC-StyleGAN trained on the LSUN-Church dataset.}
    \label{fig:church_gen_02}
\end{figure}

We provide extra image generation results from our QC-StyleGAN on the FFHQ (Fig. \ref{fig:ffhq_gen_01} and \ref{fig:ffhq_gen_02}), AFHQ-Cat (Fig. \ref{fig:cat_gen_01} and \ref{fig:cat_gen_02}), and LSUN-Church (Fig. \ref{fig:church_gen_01} and \ref{fig:church_gen_02}).

\subsection{Face image restoration}
In this section we provide additional qualitative results of the proposed face image restoration method, compared with state-of-the-art methods. We test two scenarios: image super-resolution and general image restoration.

\begin{figure}
    \setlength{\tabcolsep}{0.3pt}
    \begin{center}
    \begin{tabular}{cccc}
        Input & PULSE \cite{menon2020pulse} & HiFaceGAN \cite{yang2020hifacegan} & Real-ESRGAN  \\
        \fourimgperrow{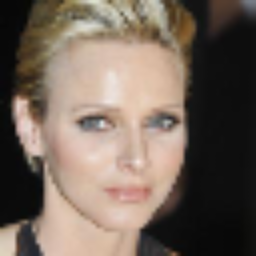} &
        \fourimgperrow{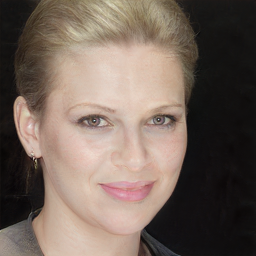} &
        \fourimgperrow{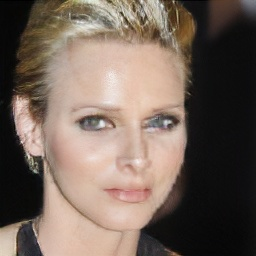} &
        \fourimgperrow{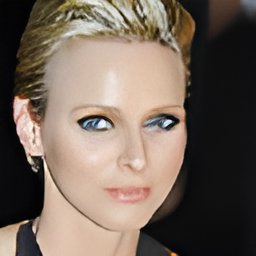}\\
        NAFNet \cite{chen2022simple} & MPRNet \cite{karras2018progressive} & Ours & Sharp \\
        \fourimgperrow{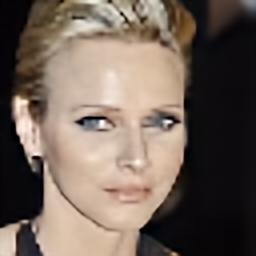} &
        \fourimgperrow{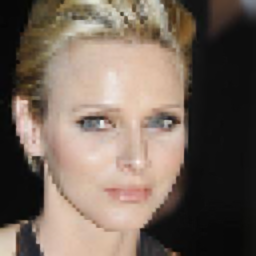} &
        \fourimgperrow{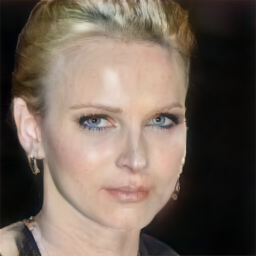} &
        \fourimgperrow{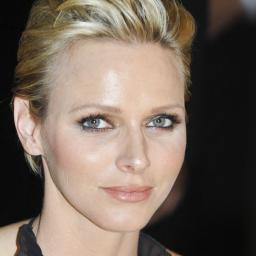}\\
    \end{tabular}
    \caption{Additional super-resolution qualitative result on the CelebA-HQ dataset}
    \label{fig:srfig01}
    \end{center}
\end{figure}

\begin{figure}
    \setlength{\tabcolsep}{0.3pt}
    \begin{center}
    \begin{tabular}{cccc}
        Input & PULSE \cite{menon2020pulse} & HiFaceGAN \cite{yang2020hifacegan} & Real-ESRGAN  \\
        \fourimgperrow{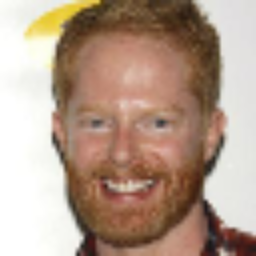} &
        \fourimgperrow{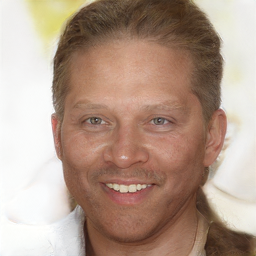} &
        \fourimgperrow{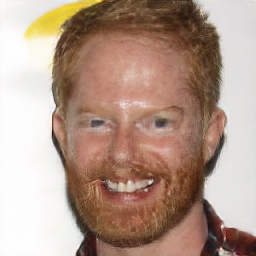} &
        \fourimgperrow{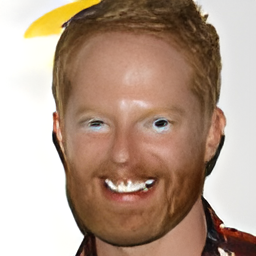}\\
        NAFNet \cite{chen2022simple} & MPRNet \cite{karras2018progressive} & Ours & Sharp \\
        \fourimgperrow{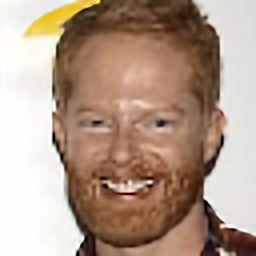} &
        \fourimgperrow{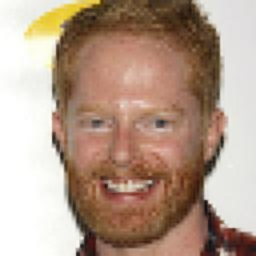} &
        \fourimgperrow{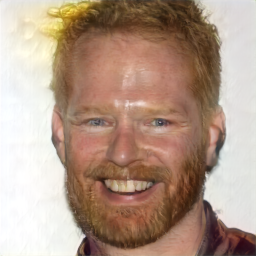} &
        \fourimgperrow{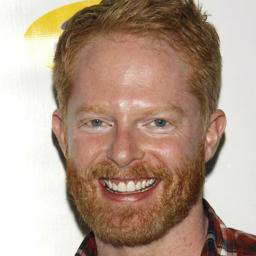}\\
    \end{tabular}
    \caption{Additional super-resolution qualitative result on the CelebA-HQ dataset}
    \label{fig:srfig02}
    \end{center}
\end{figure}

\begin{figure}
    \setlength{\tabcolsep}{0.3pt}
    \begin{center}
    \begin{tabular}{cccc}
        Input & PULSE \cite{menon2020pulse} & HiFaceGAN \cite{yang2020hifacegan} & Real-ESRGAN  \\
        \fourimgperrow{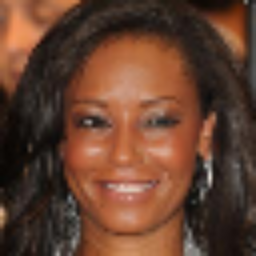} &
        \fourimgperrow{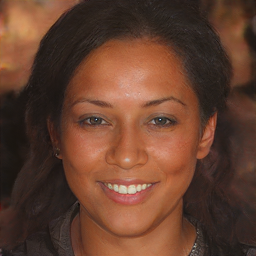} &
        \fourimgperrow{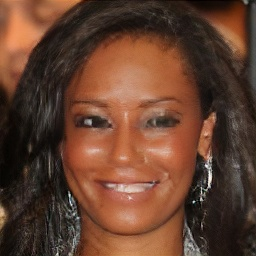} &
        \fourimgperrow{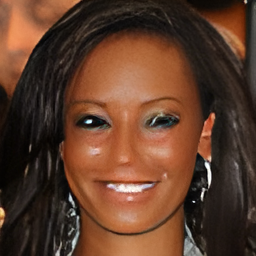}\\
        NAFNet \cite{chen2022simple} & MPRNet \cite{karras2018progressive} & Ours & Sharp \\
        \fourimgperrow{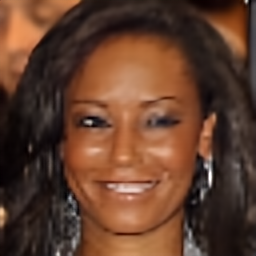} &
        \fourimgperrow{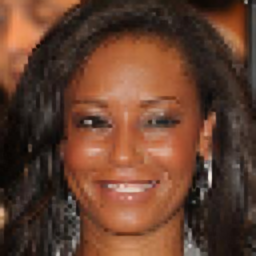} &
        \fourimgperrow{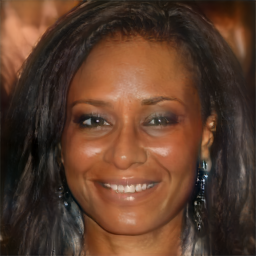} &
        \fourimgperrow{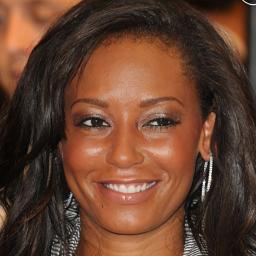}\\
    \end{tabular}
    \caption{Additional super-resolution qualitative result on the CelebA-HQ dataset}
    \label{fig:srfig03}
    \end{center}
\end{figure}

\begin{figure}
    \setlength{\tabcolsep}{0.3pt}
    \begin{center}
    \begin{tabular}{cccc}
        Input & PULSE \cite{menon2020pulse} & HiFaceGAN \cite{yang2020hifacegan} & Real-ESRGAN  \\
        \fourimgperrow{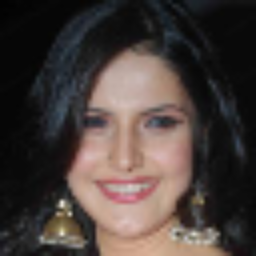} &
        \fourimgperrow{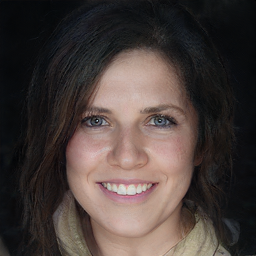} &
        \fourimgperrow{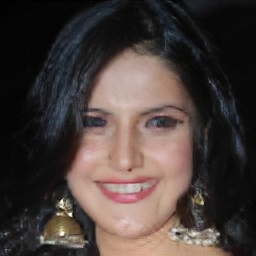} &
        \fourimgperrow{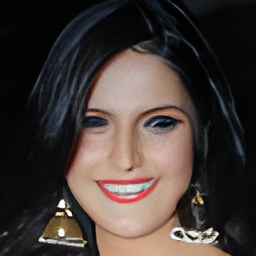}\\
        NAFNet \cite{chen2022simple} & MPRNet \cite{karras2018progressive} & Ours & Sharp \\
        \fourimgperrow{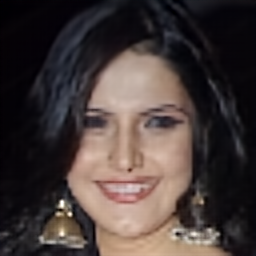} &
        \fourimgperrow{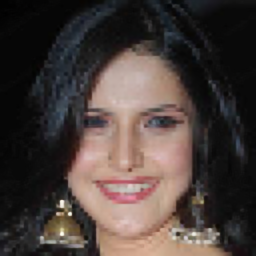} &
        \fourimgperrow{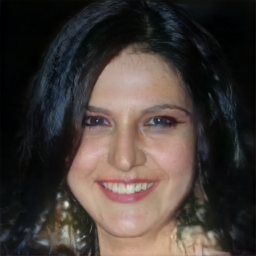} &
        \fourimgperrow{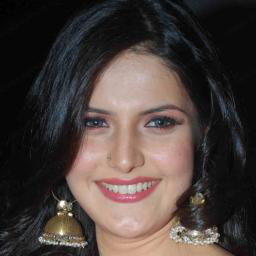}\\
    \end{tabular}
    \caption{Additional super-resolution qualitative result on the CelebA-HQ dataset}
    \label{fig:srfig04}
    \end{center}
\end{figure}

\begin{figure}
    \setlength{\tabcolsep}{0.3pt}
    \begin{center}
    \begin{tabular}{cccc}
        Input & PULSE \cite{menon2020pulse} & HiFaceGAN \cite{yang2020hifacegan} & Real-ESRGAN  \\
        \fourimgperrow{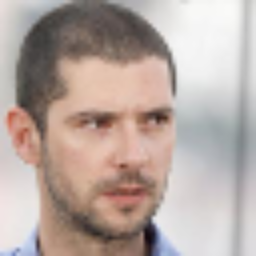} &
        \fourimgperrow{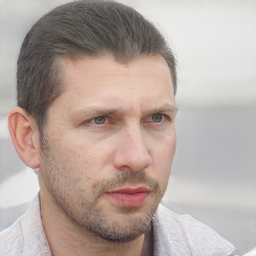} &
        \fourimgperrow{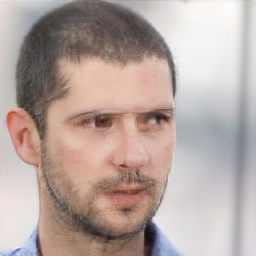} &
        \fourimgperrow{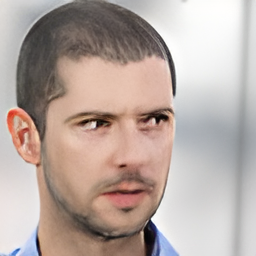}\\
        NAFNet \cite{chen2022simple} & MPRNet \cite{karras2018progressive} & Ours & Sharp \\
        \fourimgperrow{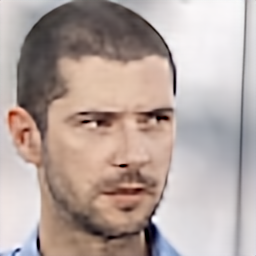} &
        \fourimgperrow{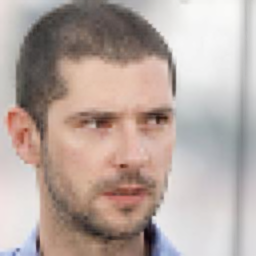} &
        \fourimgperrow{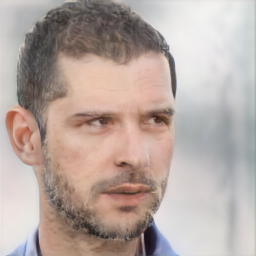} &
        \fourimgperrow{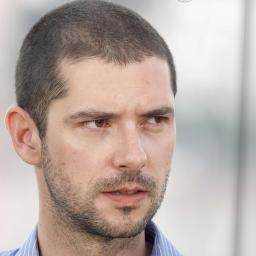}\\
    \end{tabular}
    \caption{Additional super-resolution qualitative result on the CelebA-HQ dataset}
    \label{fig:srfig05}
    \end{center}
\end{figure}

\begin{figure}
    \setlength{\tabcolsep}{0.3pt}
    \begin{center}
    \begin{tabular}{cccc}
        Input & PULSE \cite{menon2020pulse} & HiFaceGAN     \cite{yang2020hifacegan} & Real-ESRGAN  \\
        \fourimgperrow{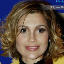} &
        \fourimgperrow{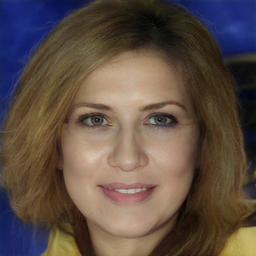} &
        \fourimgperrow{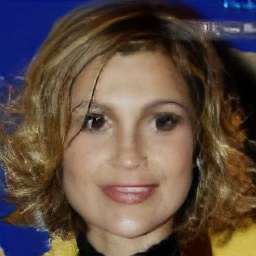} &
        \fourimgperrow{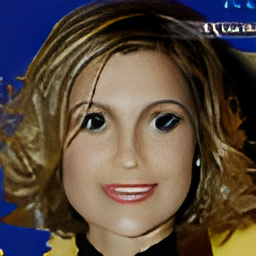}\\
        NAFNet \cite{chen2022simple} & MPRNet \cite{karras2018progressive} & Ours & Sharp \\
        \fourimgperrow{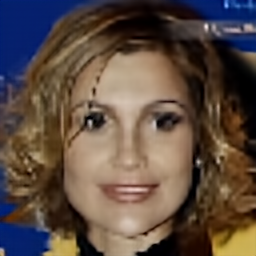} &
        \fourimgperrow{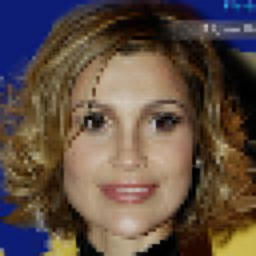} &
        \fourimgperrow{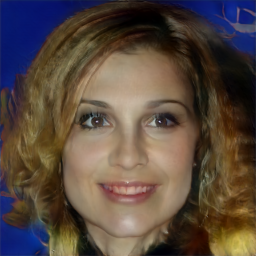} &
        \fourimgperrow{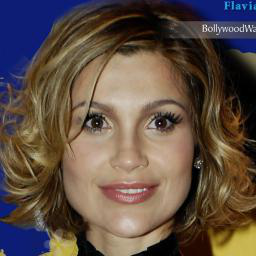}\\
    \end{tabular}
    \caption{Additional super-resolution qualitative result on the CelebA-HQ dataset}
    \label{fig:srfig06}
    \end{center}
\end{figure}

With the image super-resolution task, each sharp image is downsampled 4 times, using bilinear interpolation, to get the low-resolution input images. The state-of-the-art baselines include PULSE \cite{menon2020pulse}, HiFaceGAN \cite{yang2020hifacegan}, Real-ESRGAN \cite{wang2021realesrgan}, NAFNet \cite{chen2022simple}, and MPRNet \cite{zamir2021multi}. Qualitative results are illustrated in Fig.  \ref{fig:srfig01}, \ref{fig:srfig02}, \ref{fig:srfig03}, \ref{fig:srfig04}, \ref{fig:srfig05}, and \ref{fig:srfig06}.

\begin{figure}
    \setlength{\tabcolsep}{0.3pt}
    \begin{center}
    \begin{tabular}{ccc}
        Input &  HiFaceGAN \cite{yang2020hifacegan} & NAFNet \cite{chen2022simple}  \\
        \threeimgperrow{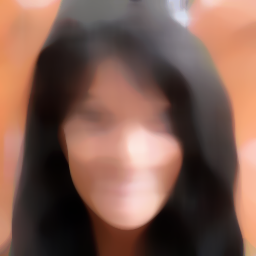} &
        \threeimgperrow{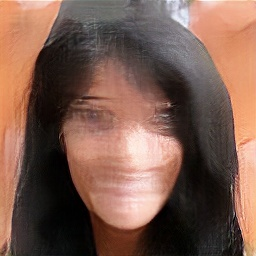} &
        \threeimgperrow{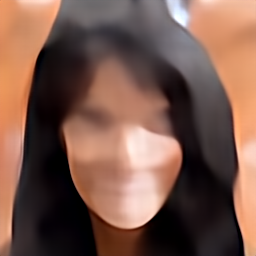}\\
        MPRNet \cite{karras2018progressive} & Ours & Sharp \\
        \threeimgperrow{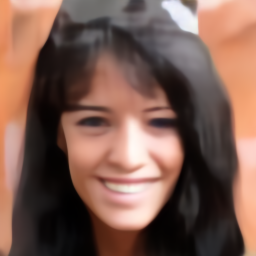} &
        \threeimgperrow{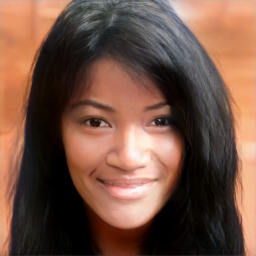} &
        \threeimgperrow{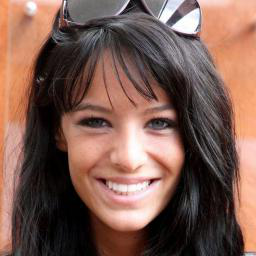}\\
    \end{tabular}
    \caption{Additional image deblurring qualitative result on the CelebA-HQ dataset}
    \label{fig:blfig01}
    \end{center}
\end{figure}

\begin{figure}
    \setlength{\tabcolsep}{0.3pt}
    \begin{center}
    \begin{tabular}{ccc}
        Input &  HiFaceGAN \cite{yang2020hifacegan} & NAFNet \cite{chen2022simple}  \\
        \threeimgperrow{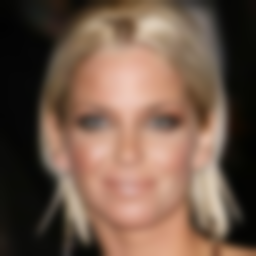} &
        \threeimgperrow{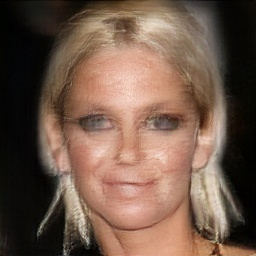} &
        \threeimgperrow{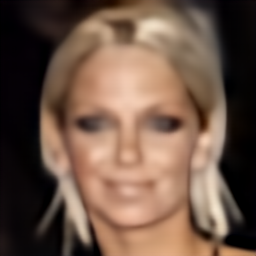} \\
        MPRNet \cite{karras2018progressive} & Ours & Sharp \\
        \threeimgperrow{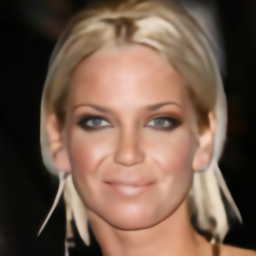} &
        \threeimgperrow{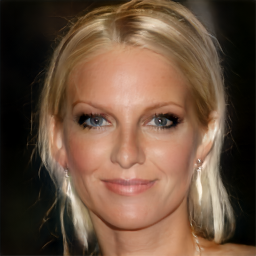} &
        \threeimgperrow{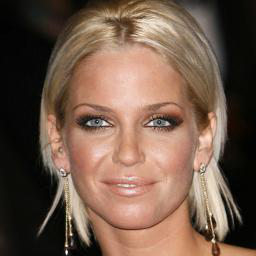}\\
    \end{tabular}
    \caption{Additional image deblurring qualitative result on the CelebA-HQ dataset}
    \label{fig:blfig02}
    \end{center}
\end{figure}

\begin{figure}
    \setlength{\tabcolsep}{0.3pt}
    \begin{center}
    \begin{tabular}{ccc}
        Input &  HiFaceGAN \cite{yang2020hifacegan} & NAFNet \cite{chen2022simple}  \\
        \threeimgperrow{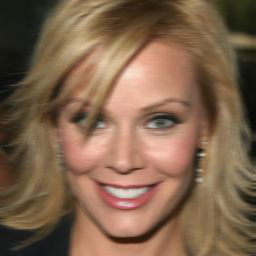} &
        \threeimgperrow{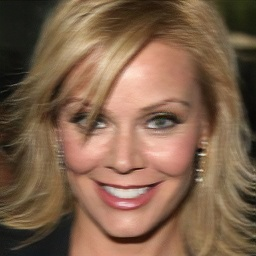} &
        \threeimgperrow{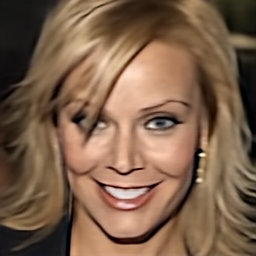}\\
        MPRNet \cite{karras2018progressive} & Ours & Sharp \\
        \threeimgperrow{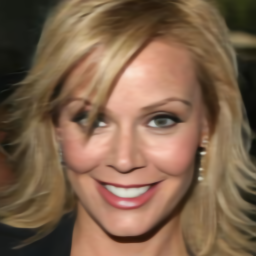} &
        \threeimgperrow{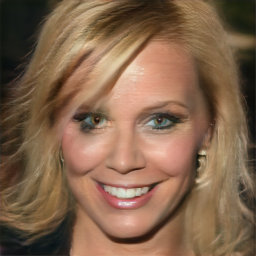} &
        \threeimgperrow{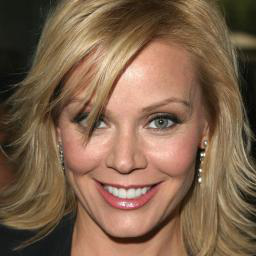}\\
    \end{tabular}
    \caption{Additional image deblurring qualitative result on the CelebA-HQ dataset}
    \label{fig:blfig03}
    \end{center}
\end{figure}

\begin{figure}
    \setlength{\tabcolsep}{0.3pt}
    \begin{center}
    \begin{tabular}{ccc}
        Input &  HiFaceGAN \cite{yang2020hifacegan} & NAFNet \cite{chen2022simple}  \\
        \threeimgperrow{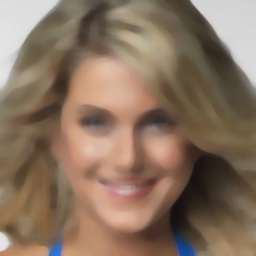} &
        \threeimgperrow{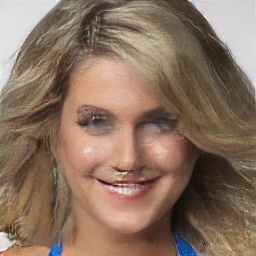} &
        \threeimgperrow{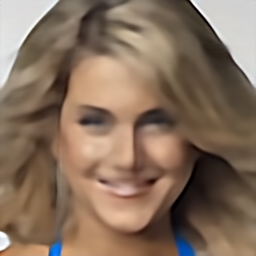} \\
        MPRNet \cite{karras2018progressive} & Ours & Sharp \\
        \threeimgperrow{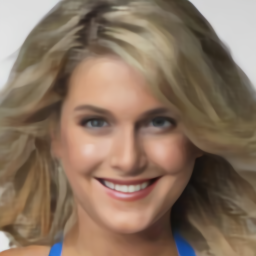} &
        \threeimgperrow{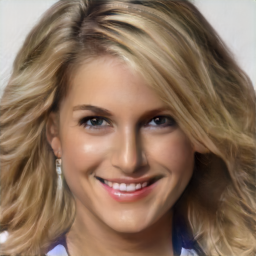} &
        \threeimgperrow{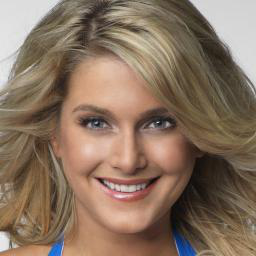}\\
    \end{tabular}
    \caption{Additional image deblurring qualitative result on the CelebA-HQ dataset}
    \label{fig:blfig04}
    \end{center}
\end{figure}

\begin{figure}
    \setlength{\tabcolsep}{0.3pt}
    \begin{center}
    \begin{tabular}{ccc}
        Input &  HiFaceGAN \cite{yang2020hifacegan} & NAFNet \cite{chen2022simple}  \\
        \threeimgperrow{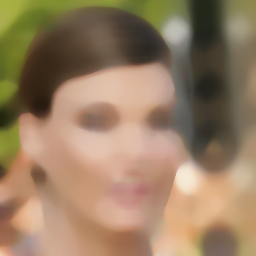} &
        \threeimgperrow{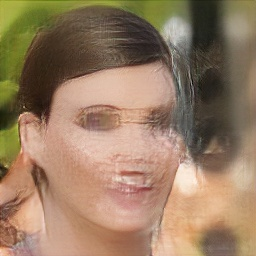} &
        \threeimgperrow{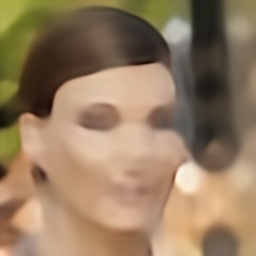} \\
        MPRNet \cite{karras2018progressive} & Ours & Sharp \\
        \threeimgperrow{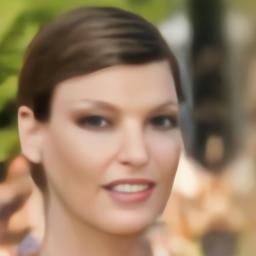} &
        \threeimgperrow{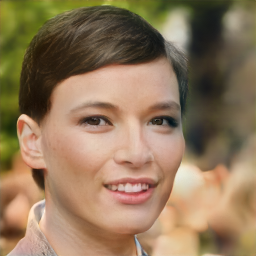} &
        \threeimgperrow{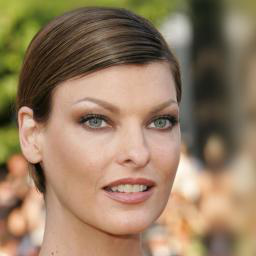}\\
    \end{tabular}
    \caption{Additional image deblurring qualitative result on the CelebA-HQ dataset}
    \label{fig:blfig05}
    \end{center}
\end{figure}

\begin{figure}
    \setlength{\tabcolsep}{0.3pt}
    \begin{center}
    \begin{tabular}{ccc}
        Input &  HiFaceGAN \cite{yang2020hifacegan} & NAFNet \cite{chen2022simple}  \\
        \threeimgperrow{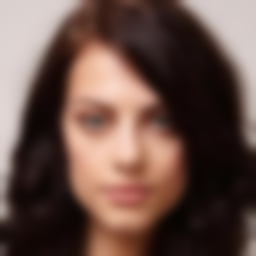} &
        \threeimgperrow{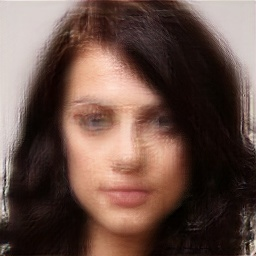} &
        \threeimgperrow{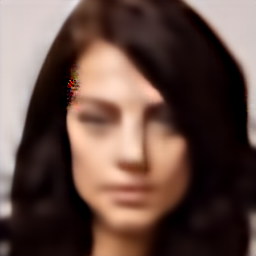} \\
        MPRNet \cite{karras2018progressive} & Ours & Sharp \\
        \threeimgperrow{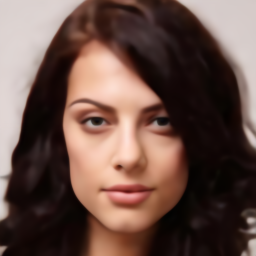} &
        \threeimgperrow{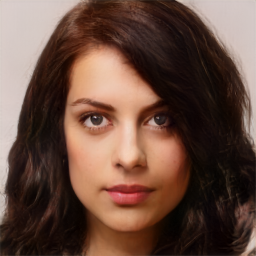} &
        \threeimgperrow{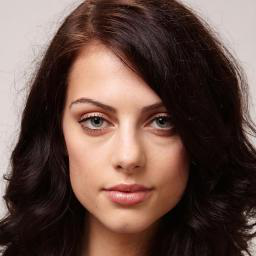}\\
    \end{tabular}
    \caption{Additional image deblurring qualitative result on the CelebA-HQ dataset}
    \label{fig:blfig06}
    \end{center}
\end{figure}

\begin{figure}
    \setlength{\tabcolsep}{0.3pt}
    \begin{center}
    \begin{tabular}{ccc}
        Input &  HiFaceGAN \cite{yang2020hifacegan} & NAFNet \cite{chen2022simple}  \\
        \threeimgperrow{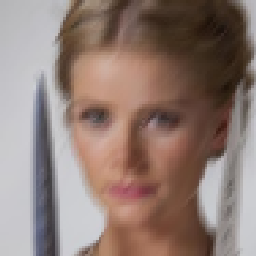} &
        \threeimgperrow{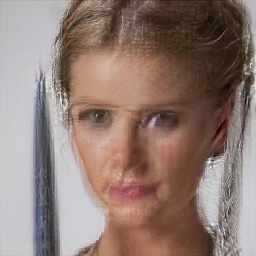} &
        \threeimgperrow{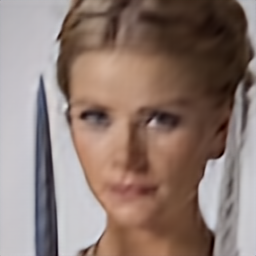} \\
        MPRNet \cite{karras2018progressive} & Ours & Sharp \\
        \threeimgperrow{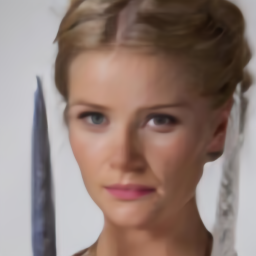} &
        \threeimgperrow{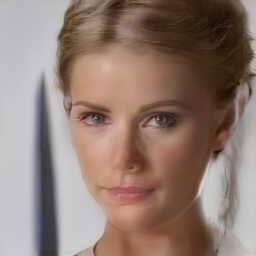} &
        \threeimgperrow{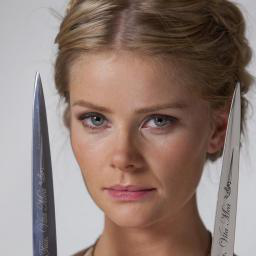}\\
    \end{tabular}
    \caption{Additional multi-degradation image restoration result on the CelebA-HQ dataset}
    \label{fig:mdfig01}
    \end{center}
\end{figure}

\begin{figure}
    \setlength{\tabcolsep}{0.3pt}
    \begin{center}
    \begin{tabular}{ccc}
        Input &  HiFaceGAN \cite{yang2020hifacegan} & NAFNet \cite{chen2022simple}  \\
        \threeimgperrow{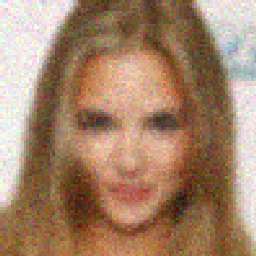} &
        \threeimgperrow{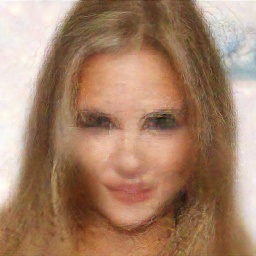} &
        \threeimgperrow{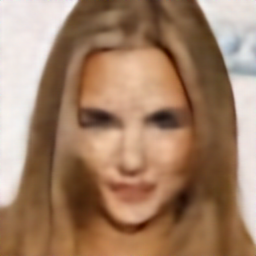} \\
        MPRNet \cite{karras2018progressive} & Ours & Sharp \\
        \threeimgperrow{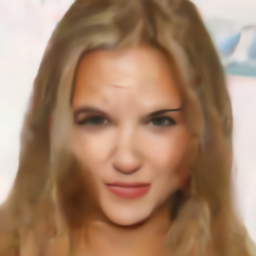} &
        \threeimgperrow{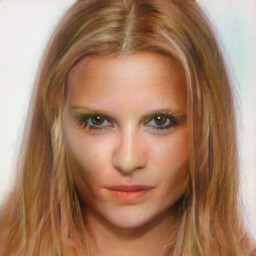} &
        \threeimgperrow{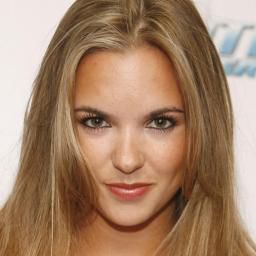}\\
    \end{tabular}
    \caption{Additional multi-degradation image restoration result on the CelebA-HQ dataset}
    \label{fig:mdfig02}
    \end{center}
\end{figure}

\begin{figure}
    \setlength{\tabcolsep}{0.3pt}
    \begin{center}
    \begin{tabular}{ccc}
        Input &  HiFaceGAN \cite{yang2020hifacegan} & NAFNet \cite{chen2022simple}  \\
        \threeimgperrow{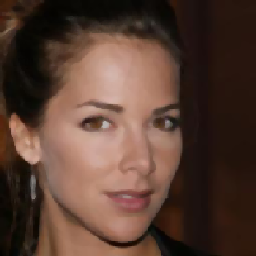} &
        \threeimgperrow{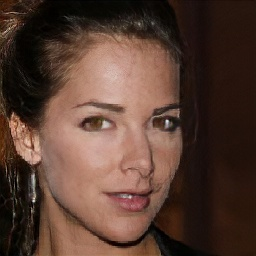} &
        \threeimgperrow{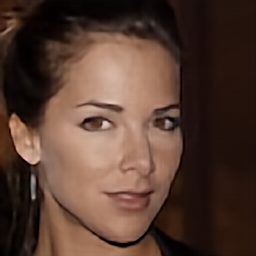} \\
        MPRNet \cite{karras2018progressive} & Ours & Sharp \\
        \threeimgperrow{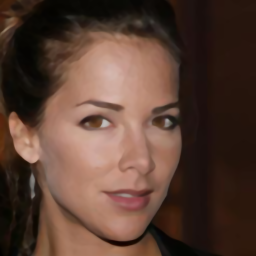} &
        \threeimgperrow{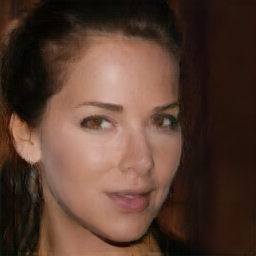} &
        \threeimgperrow{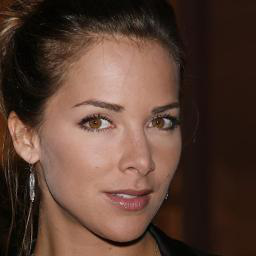}\\
    \end{tabular}
    \caption{Additional multi-degradation image restoration result on the CelebA-HQ dataset}
    \label{fig:mdfig03}
    \end{center}
\end{figure}

\begin{figure}
    \setlength{\tabcolsep}{0.3pt}
    \begin{center}
    \begin{tabular}{ccc}
        Input &  HiFaceGAN \cite{yang2020hifacegan} & NAFNet \cite{chen2022simple}  \\
        \threeimgperrow{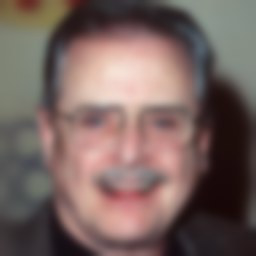} &
        \threeimgperrow{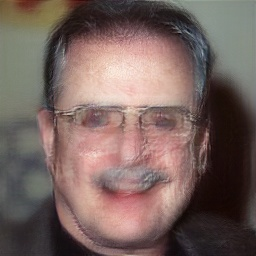} &
        \threeimgperrow{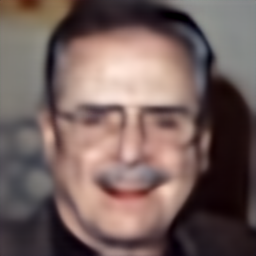} \\
        MPRNet \cite{karras2018progressive} & Ours & Sharp \\
        \threeimgperrow{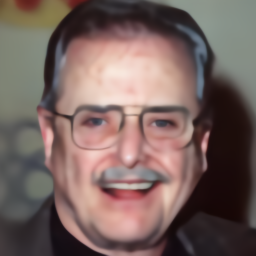} &
        \threeimgperrow{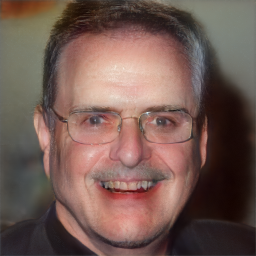} &
        \threeimgperrow{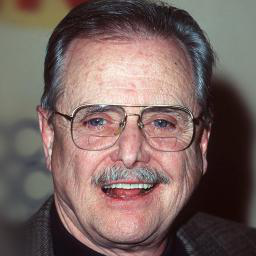}\\
    \end{tabular}
    \caption{Additional multi-degradation image restoration result on the CelebA-HQ dataset}
    \label{fig:mdfig04}
    \end{center}
\end{figure}

\begin{figure}
    \setlength{\tabcolsep}{0.3pt}
    \begin{center}
    \begin{tabular}{ccc}
        Input &  HiFaceGAN \cite{yang2020hifacegan} & NAFNet \cite{chen2022simple}  \\
        \threeimgperrow{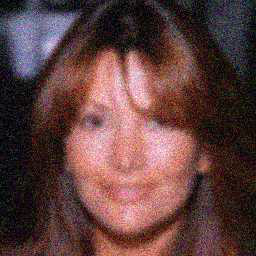} &
        \threeimgperrow{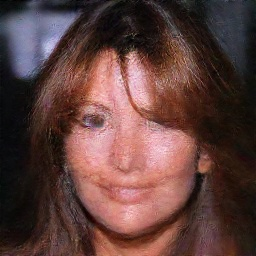} &
        \threeimgperrow{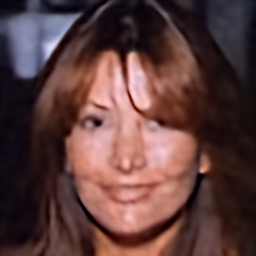} \\
        MPRNet \cite{karras2018progressive} & Ours & Sharp \\
        \threeimgperrow{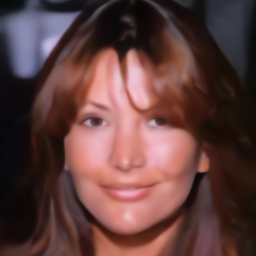} &
        \threeimgperrow{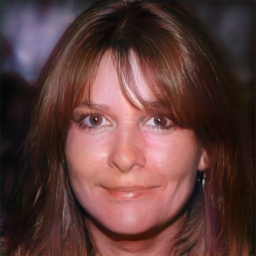} &
        \threeimgperrow{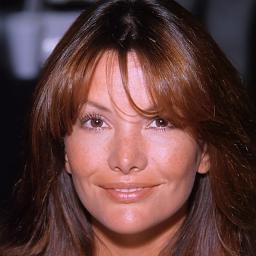}\\
    \end{tabular}
    \caption{Additional multi-degradation image restoration result on the CelebA-HQ dataset}
    \label{fig:mdfig05}
    \end{center}
\end{figure}

\begin{figure}
    \setlength{\tabcolsep}{0.3pt}
    \begin{center}
    \begin{tabular}{ccc}
        Input &  HiFaceGAN \cite{yang2020hifacegan} & NAFNet \cite{chen2022simple}  \\
        \threeimgperrow{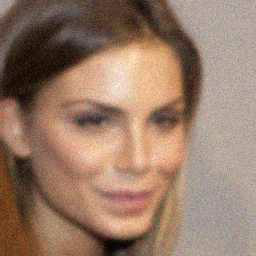} &
        \threeimgperrow{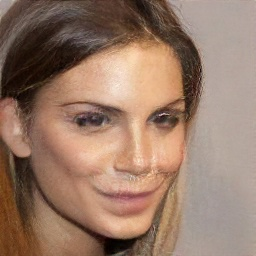} &
        \threeimgperrow{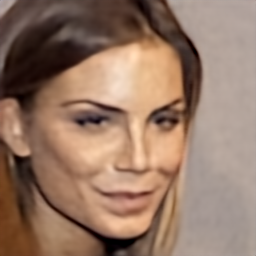} \\
        MPRNet \cite{karras2018progressive} & Ours & Sharp \\
        \threeimgperrow{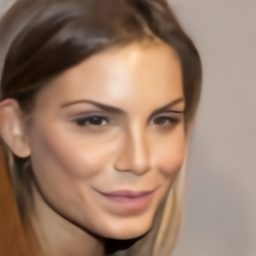} &
        \threeimgperrow{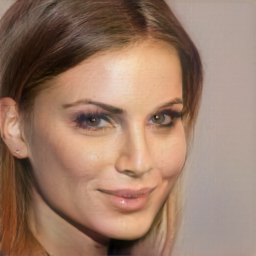} &
        \threeimgperrow{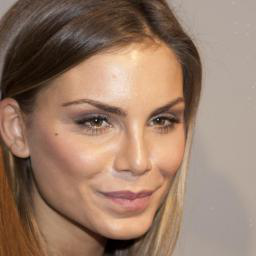}\\
    \end{tabular}
    \caption{Additional multi-degradation image restoration result on the CelebA-HQ dataset}
    \label{fig:mdfig06}
    \end{center}
\end{figure}

With general image super-resolution, the state-of-the-art baselines include HiFaceGAN \cite{yang2020hifacegan}, NAFNet \cite{chen2022simple}, and MPRNet \cite{zamir2021multi}. Qualitative results are illustrated in Fig. \ref{fig:blfig01}, \ref{fig:blfig02}, \ref{fig:blfig03}, \ref{fig:blfig04}, \ref{fig:blfig05},  \ref{fig:blfig06}, \ref{fig:mdfig01}, \ref{fig:mdfig02}, \ref{fig:mdfig03}, \ref{fig:mdfig04}, \ref{fig:mdfig05}, and \ref{fig:mdfig06}.

\subsection{Face editing}
\begin{figure}[ht]
    \setlength{\tabcolsep}{0.3pt}
    \begin{center}
    \begin{tabular}{cccc}
        \fourimgperrow{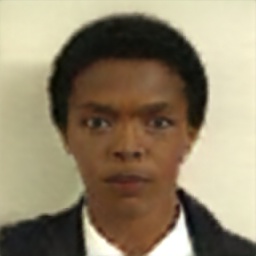} &
        \fourimgperrow{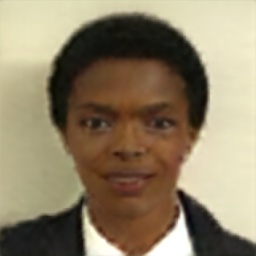} &
        \fourimgperrow{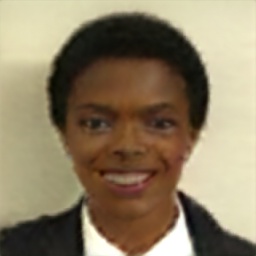} &
        \fourimgperrow{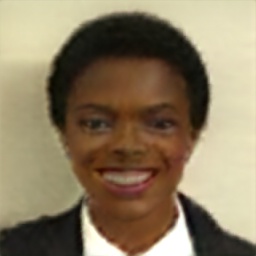}\\
        \fourimgperrow{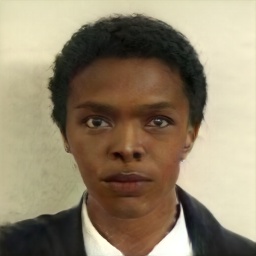} &
        \fourimgperrow{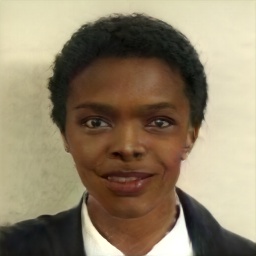} &
        \fourimgperrow{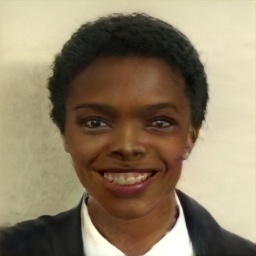} &
        \fourimgperrow{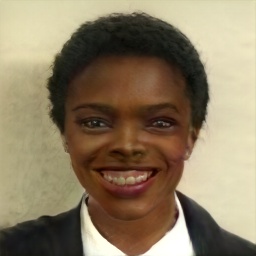}\\
        \fourimgperrow{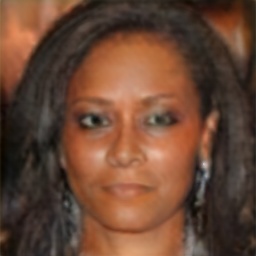} &
        \fourimgperrow{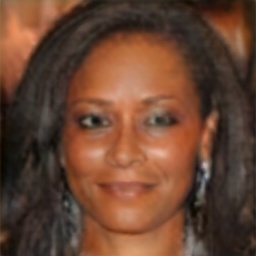} &
        \fourimgperrow{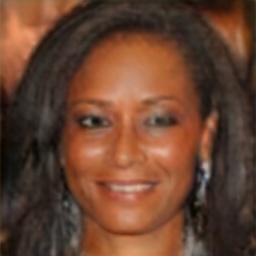} &
        \fourimgperrow{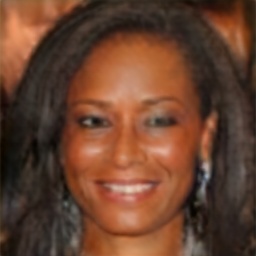}\\
        \fourimgperrow{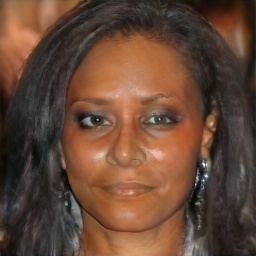} &
        \fourimgperrow{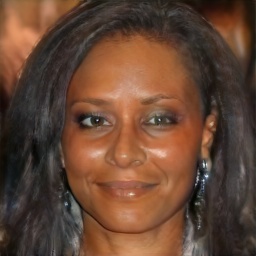} &
        \fourimgperrow{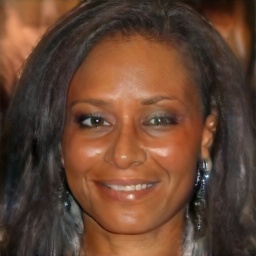} &
        \fourimgperrow{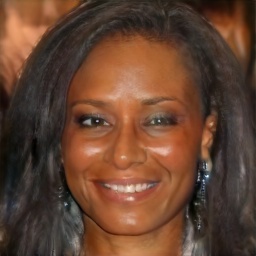}\\
        \fourimgperrow{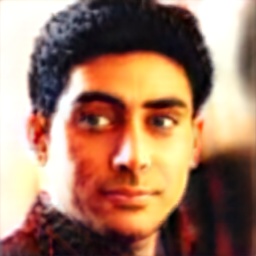} &
        \fourimgperrow{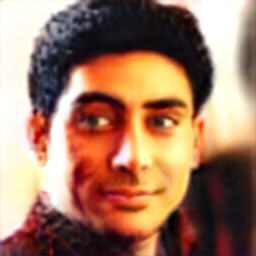} &
        \fourimgperrow{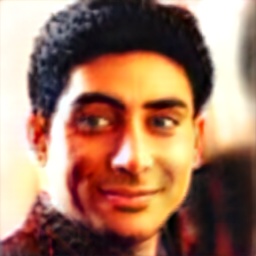} &
        \fourimgperrow{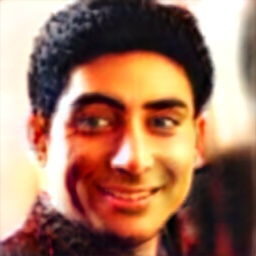}\\
        \fourimgperrow{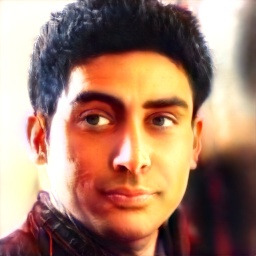} &
        \fourimgperrow{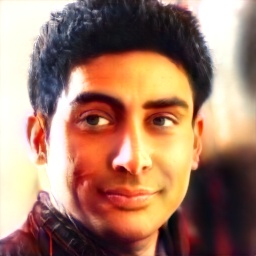} &
        \fourimgperrow{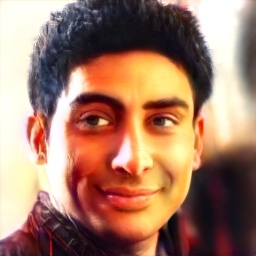} &
        \fourimgperrow{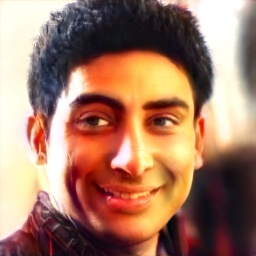}\\
    \end{tabular}
    \caption{\textbf{GAN inversion and editing}. Each example is shown in two rows: For a given low-resolution image ($1^{st}$ image), we inverse it and apply the ``smiling'' editing direction on the latent code. The first row shows the edited images in the degraded domain, while the second row contains the edited images in the sharp domain.}
    \label{fig:edit_smile_sr}
    \end{center}
\end{figure}

\begin{figure}[ht]
    \setlength{\tabcolsep}{0.3pt}
    \begin{center}
    \begin{tabular}{cccc}
        \fourimgperrow{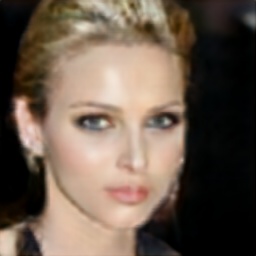} &
        \fourimgperrow{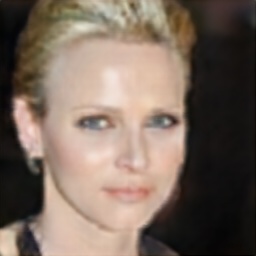} &
        \fourimgperrow{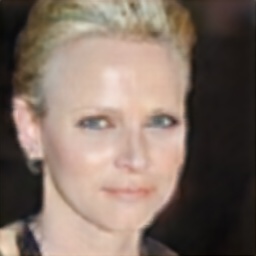} &
        \fourimgperrow{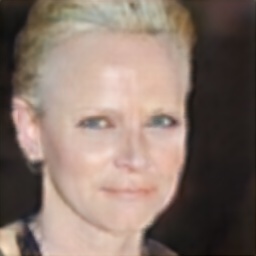}\\
        \fourimgperrow{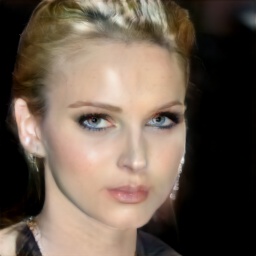} &
        \fourimgperrow{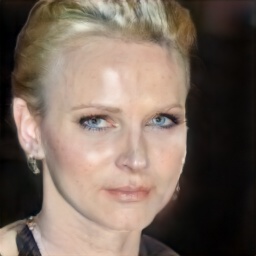} &
        \fourimgperrow{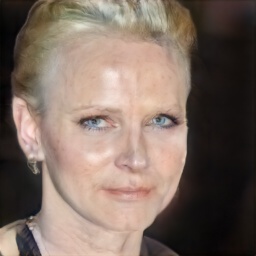} &
        \fourimgperrow{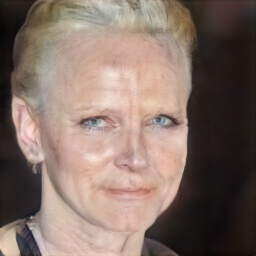}\\
        \fourimgperrow{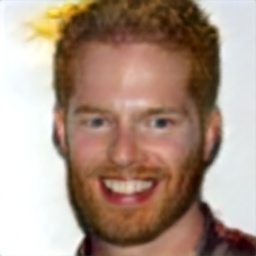} &
        \fourimgperrow{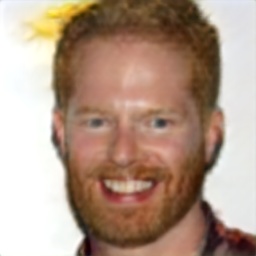} &
        \fourimgperrow{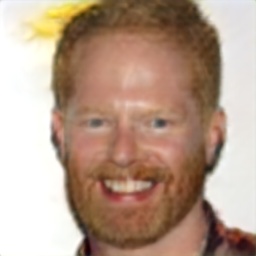} &
        \fourimgperrow{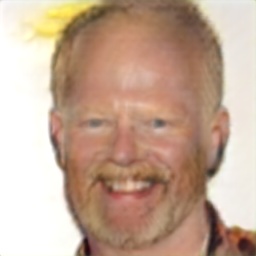}\\
        \fourimgperrow{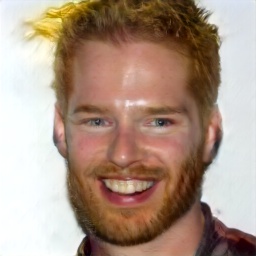} &
        \fourimgperrow{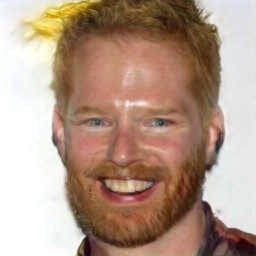} &
        \fourimgperrow{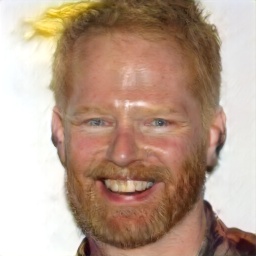} &
        \fourimgperrow{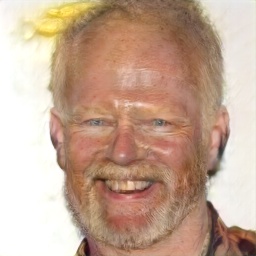}\\
        \fourimgperrow{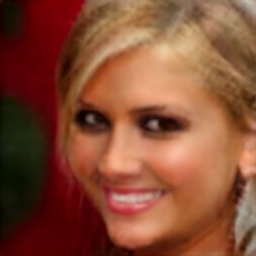} &
        \fourimgperrow{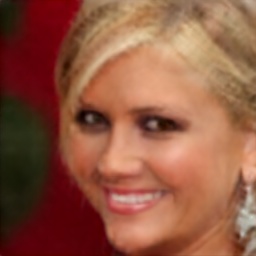} &
        \fourimgperrow{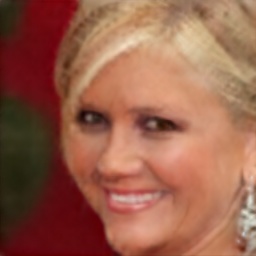} &
        \fourimgperrow{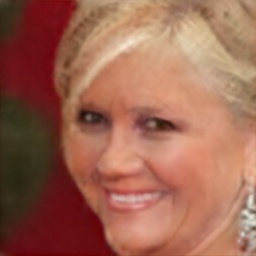}\\
        \fourimgperrow{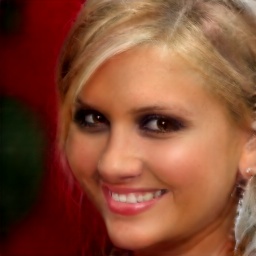} &
        \fourimgperrow{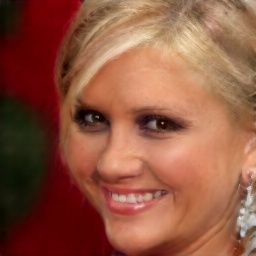} &
        \fourimgperrow{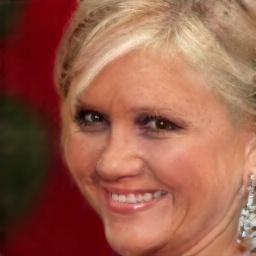} &
        \fourimgperrow{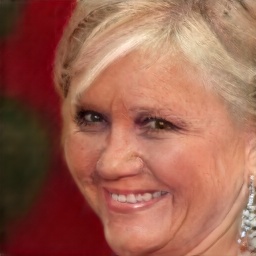}\\
    \end{tabular}
    \caption{\textbf{GAN inversion and editing}. Each example is shown in two rows: For a given low-resolution image ($1^{st}$ image), we inverse it and apply the ``aging'' editing direction on the latent code. The first row shows the edited images in the degraded domain, while the second row contains the edited images in the sharp domain.}
    \label{fig:edit_age_sr}
    \end{center}
\end{figure}

In this section, we provide extra image editing results on facial images. We use InterfaceGAN \cite{shen2020interfacegan} to find the editing direction for ``smiling'' and ``aging''. Given a degraded input image, we first perform image inversion, then apply the learned editing on the optimized latent code to change the expression and age of the input face. Qualitative results are given in \Fref{fig:edit_smile_sr} and \ref{fig:edit_age_sr}.


\subsection{Degradation synthesis}
\begin{figure}[t]
    \centering
    \includegraphics[width=\textwidth]{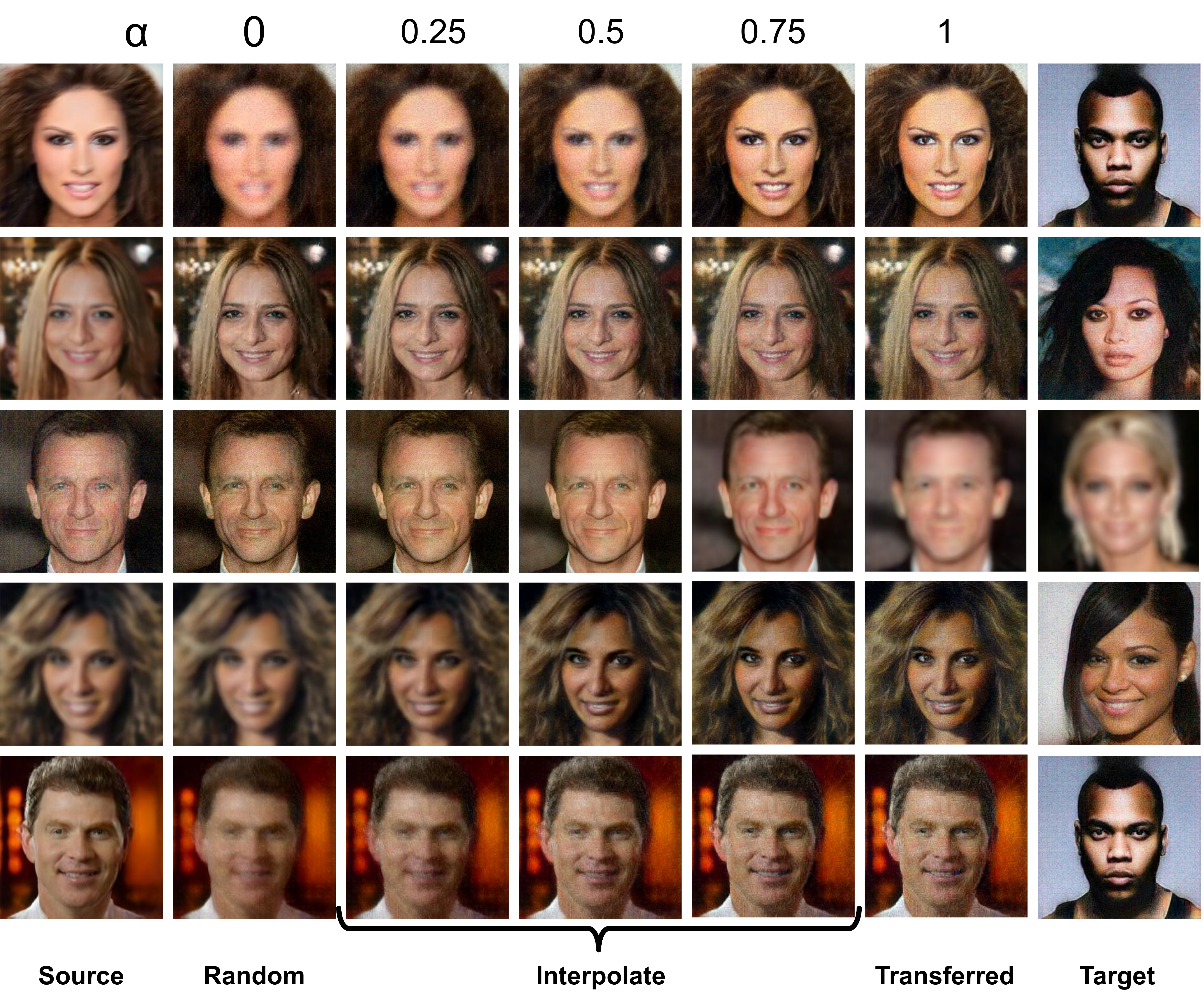}
    \caption{\textbf{Degradation synthesis.} In each row, from a source degraded image ($1^{st}$ column), we change its image degradation to a novel random one ($2^{nd}$ column) or copy the degradation from a reference image ($6^{th}$ column). We can also smoothly interpolate image degradations in-between the ones above, using an interpolation factor $\alpha \in [0, 1]$ ($3-5^{th}$ columns).} 
    \label{fig:sup_deg_syn}
    \vspace{-2mm}
\end{figure}

We provide extra image degradation synthesis  results in \Fref{fig:sup_deg_syn}.

\subsection{Image Restoration with Small Degradations}
\begin{figure}[ht]
    \centering
    \includegraphics[width=\textwidth]{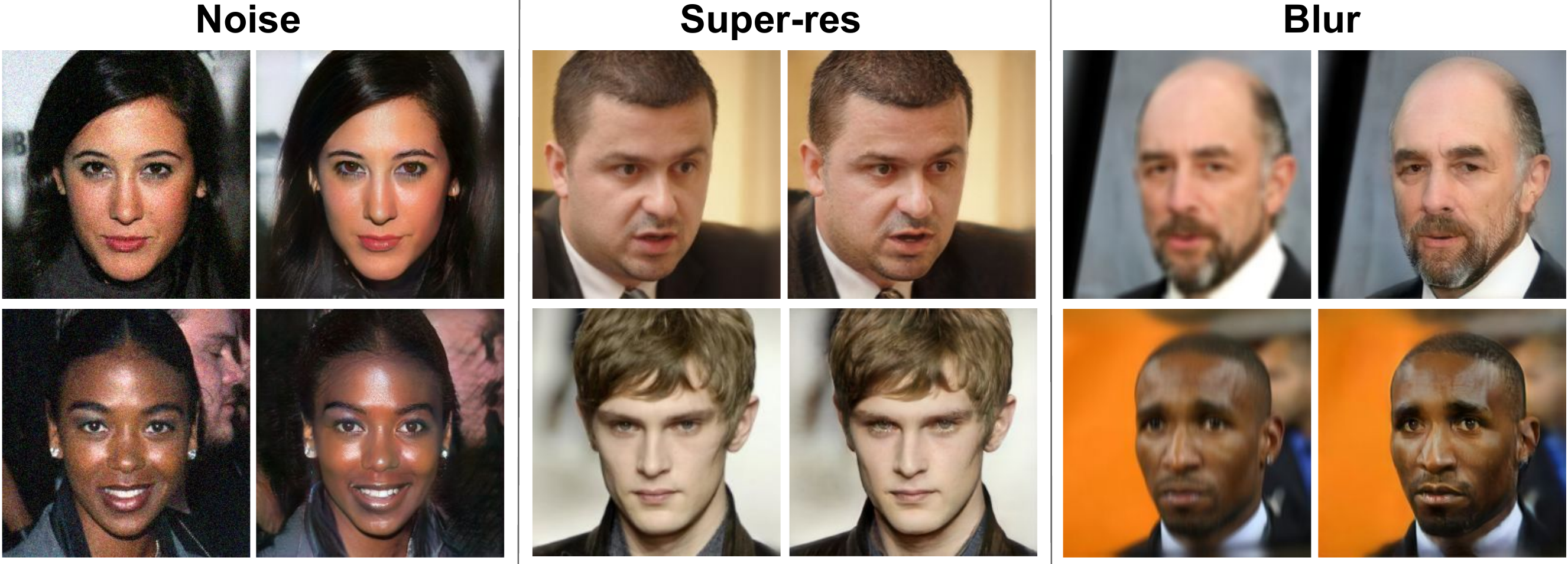}
    \caption{\textbf{Image restoration with small degradations.} For each degradation, we provide two examples from the CelebA-HQ dataset. Each example consists of the degraded input (left) and the restored image using our QC-StyleGAN. Best view in zoom.}
    \label{fig:small_deg}
\end{figure}

Our QC-StyleGAN models were trained to handle image degradations at various degrees, unlike many deep-learning-based image restoration techniques. We provide some image restoration results on CelebA-HQ images under small degradations, using our QC-StyleGAN model trained on the FFHQ dataset, in Fig. \ref{fig:small_deg}. As can be seen, these images are still recovered effectively.

\end{document}